\documentclass{article}

\usepackage{PRIMEarxiv}

\usepackage[utf8]{inputenc}
\usepackage[T1]{fontenc}
\usepackage{microtype}

\usepackage{amsmath}
\usepackage{amssymb}
\usepackage{amsfonts}
\usepackage{amsthm}

\newtheorem{proposition}{Proposition}

\newtheorem{definition}{Definition}

\usepackage{enumitem}
\usepackage{booktabs}
\usepackage[dvipsnames]{xcolor}
\usepackage{graphicx}
\usepackage{placeins}

\usepackage{fancyhdr}
\usepackage{url}
\usepackage{natbib}
\setcitestyle{authoryear,round,aysep={, },yysep={, },semicolon}

\usepackage[hidelinks]{hyperref}

\makeatletter
\renewcommand\@biblabel[1]{}
\makeatother

\newcommand{\SM}[1]{Section~#1 of the Supplementary Materials}

\title{Model-Agnostic Decomposition of the Difference in Mean Response Between Two Groups via Game-Theoretic Attribution with Minimum-Switching-Distortion Coupling}

\author{
  Hyunwook Koh\textsuperscript{*} \\
  Department of Applied Mathematics and Statistics \\
  The State University of New York, Korea \\
  119 Songdo Moonhwa-ro, Yeonsu-gu, Incheon 21985, South Korea\\
  \textsuperscript{*}Corresponding author. Email: \texttt{hyunwook.koh@stonybrook.edu}}

\begin{document}

\maketitle

\begin{abstract}

The difference in mean response between two groups can arise from differences in covariate–response relationships, covariate distributions, or both. We propose a model-agnostic decomposition (MAD) of this difference into structure and composition effects. We formulate a cooperative switching game with a model-switch player and individual covariate-switch players, organized into model and covariate unions. MAD employs the Owen value, whose quotient game yields the aggregate structure and composition effects and whose internal game allocates the aggregate composition effect across covariates, thereby inheriting its fair allocation properties. Partial covariate switches create hybrid covariate vectors whose joint distributions are not determined by the group-specific covariate distributions alone. To resolve this indeterminacy, we propose a minimum-switching-distortion (MSD) coupling, formulated as an optimal-transport problem, that satisfies identity at coincidence, transposition equivariance, and coordinate equivariance and whose sparse support reduces allocation cost. Because MAD imposes no functional form, it accommodates machine-learning models that capture nonlinear and interactive relationships, yielding more accurate decompositions. Simulation studies demonstrate the effectiveness of MAD and the MSD coupling in decomposition accuracy, coupling admissibility, and computational cost. As an illustration, we decompose the difference in mean body mass index between UK and US males aged 30–70 years using microbiome data.

\end{abstract}


\section{Introduction}
\label{sec:introduction}

The difference in mean response between two groups is of interest in many areas of economics, social science, epidemiology, and biomedical research. A central question is how much of the mean-response difference is associated with differences in covariate composition and how much with differences in the covariate--response relationship. The Blinder--Oaxaca decomposition (Blinder, 1973; Oaxaca, 1973) addresses this question under linear regression models by decomposing the mean-response difference into a composition effect due to differences in covariate means and a structure effect due to differences in regression coefficients. The linearity and additivity of the model also permit straightforward covariate-specific decompositions of the composition and structure effects.\\
\indent A substantial literature has extended the Blinder--Oaxaca decomposition beyond linear regression models. Yun (2004) and Sinning et al. (2008) developed decomposition methods for nonlinear models, while Rothe (2015) further decomposed the composition effect under general nonlinear relationships into contributions from marginal covariate distributions, their dependence, and their interactions. More recently, Mao et al. (2025) developed a double-machine-learning approach for estimation and inference on Oaxaca--Blinder counterfactual means and aggregate effects.
We likewise accommodate nonlinear and interactive predictive models, including black-box machine-learning models, without requiring a coefficient-based representation. In addition, we leverage game-theoretic attribution to ensure desirable allocation properties.\\
\indent Game-theoretic attribution provides a principled way to address order dependence in decomposition problems. The Shapley value (Shapley, 1953) offers a general principle for allocating the total value of a cooperative game across its players according to their average marginal contributions and is uniquely characterized by the axioms of efficiency, symmetry, additivity, and the null-player property. Owen (1977) extended the Shapley value to games with \textit{a priori} unions, in which players are partitioned into unions, allowing contributions to be allocated first across unions and then across players within each union. Shorrocks (2013) subsequently developed a unified game-theoretic framework for decomposition analysis, including hierarchical decompositions based on grouped contributory factors, while Audoly et al. (2025) more recently extended this line of work to nonlinear economic models. By averaging marginal contributions across possible orderings, these approaches yield order-invariant, exactly additive allocations under nonlinear relationships.\\
\indent Other Shapley-based methods have also addressed related problems involving distribution shift and predictive performance. Zhang et al. (2023) used Shapley values to decompose predictive-performance differences across environments into contributions from shifts in the underlying data-generating mechanisms, while Singh et al. (2024) hierarchically decomposed predictive-performance differences across domains into distribution-shift components and their Shapley-based covariate-specific contributions. In contrast, our target is the between-group difference in mean response rather than predictive performance, with both the predictor function and the covariate distribution differing between groups.\\
\indent In this paper, we introduce a model-agnostic decomposition (MAD) for the difference in mean response between two groups. We formulate MAD as a cooperative switching game whose players are a model switch and individual covariate switches, organized into two \textit{a priori} unions: a singleton model union and a covariate union containing all covariate switches. We then employ the Owen value (Owen, 1977) for a two-stage hierarchical allocation. In the first stage, the quotient game allocates the mean-response difference across the two unions, yielding the aggregate structure and composition effects, a two-factor Shapley decomposition (Shorrocks, 2013). In the second stage, the payoff of the covariate union, which is the aggregate composition effect, is further allocated across its covariate switches, yielding covariate-specific composition effects. While the first stage is determined entirely by the predictor functions and covariate distributions at the two endpoints, the second stage requires partial covariate switches, which create hybrid covariate vectors combining some covariates from one group with the remaining covariates from the other. The distribution of such a hybrid covariate vector is not determined by the group-specific covariate distributions alone. We therefore construct all hybrid covariate vectors from a single cross-group coupling, that is, a joint distribution pairing covariate vectors while preserving each group's covariate distribution. To this end, we propose a minimum-switching-distortion (MSD) coupling, formulated as an optimal-transport problem (Villani, 2009).\\
\indent MAD offers several advantages. First, it is model-agnostic, imposing no particular functional form and thereby accommodating modern machine-learning models that can flexibly adapt to complex nonlinear and interactive relationships. Second, its two-stage hierarchical allocation inherits the fair allocation properties of the Owen value, including exact two-level adding-up, so that the aggregate effects sum to the total difference and the covariate-specific composition effects sum to the aggregate composition effect. Third, its aggregate effects are invariant to the coupling, and under linear predictor functions, MAD reduces to the classical Blinder--Oaxaca decomposition with midpoint reference coefficients (Reimers, 1983). Finally, the MSD coupling is admissible, reducing to the identity coupling when the two covariate distributions coincide and remaining equivariant to group-label exchange and common covariate relabeling, and it has sparse support that reduces downstream allocation cost relative to alternative couplings.\\
\indent We evaluate MAD through simulation studies, whose scenarios vary in the functional form of the covariate--response relationship, including nonlinearity and covariate interactions. For each scenario, we consider predictive models ranging from traditional linear regression to modern machine-learning models, namely gradient boosting machines (Friedman, 2001) and deep neural networks (LeCun et al., 2015), and assess the recovery of population structure and composition effects. The results show that MAD with machine-learning models yields more accurate decompositions than MAD with linear regression under nonlinear and interactive relationships, while linear regression remains competitive under linear additive relationships. We also compare the MSD coupling with two alternatives: independence and entropically regularized (Cuturi, 2013) couplings. The results show that, among the three couplings, only the MSD coupling satisfies identity at coincidence, so partial covariate switches leave the internal-game coalition value unchanged when the covariate distributions coincide. Moreover, its sparse support substantially reduces allocation cost relative to the alternatives.\\
\indent We illustrate MAD using large-scale human microbiome data from the American Gut Project (McDonald et al., 2018) to decompose the difference in mean body mass index (BMI) between UK and US males aged 30--70 years into aggregate structure and composition effects and further allocate the aggregate composition effect across microbial genera. We discuss the interpretation of the genus-specific composition effects and their implications.

\section{Methods}
\label{sec:methods}

We first introduce notation and models, and review the classical Blinder--Oaxaca decomposition (Blinder, 1973; Oaxaca, 1973) under linear regression. We then formulate the model-agnostic decomposition (MAD) using a cooperative switching game and the Owen value (Owen, 1977), introduce the minimum-switching-distortion (MSD) coupling for hybrid covariate vectors, establish their theoretical properties, and describe computational procedures.

\subsection{Notation and Models}

Suppose that there are two groups, $A$ and $B$. For each group $g\in\{A,B\}$, we let $Y_g\in\mathbb{R}$ denote a response with finite mean $\mu_{Y_g}=\mathbb{E}(Y_g)$ and finite variance $\sigma_{Y_g}^2=\operatorname{Var}(Y_g)$, and let $X_g=(X_{g1},\dots,X_{gp})^\top\in\mathbb{R}^p$ denote a vector of $p$ covariates with joint distribution $P_g$, mean vector $\mu_{X_g}=\mathbb{E}(X_g)=(\mu_{X_{g1}},\dots,\mu_{X_{gp}})^\top$, and covariance matrix $\Sigma_{X_g}=\operatorname{Var}(X_g)=\mathbb{E}\{(X_g-\mu_{X_g})(X_g-\mu_{X_g})^\top\}$, where $\mu_{X_{gj}}=\mathbb{E}(X_{gj})$ and $(\Sigma_{X_g})_{jj'}=\operatorname{Cov}(X_{gj},X_{gj'})$ are finite for $j=1,\dots,p$ and $j'=1,\dots,p$.\\
\indent We consider the model
\begin{equation*}
Y_g=f_g(X_g)+\epsilon_g,
\end{equation*}
where $\epsilon_g$ is an error term satisfying $\mathbb{E}(\epsilon_g\mid X_g)=0$, and $f_g$ may be nonlinear and may include covariate interactions. The conditional and marginal means are then
\begin{equation*}
\mu_{Y_g\mid X_g}=\mathbb{E}(Y_g\mid X_g)=f_g(X_g),
\qquad
\mu_{Y_g}=\mathbb{E}_{X_g}\{f_g(X_g)\}.
\end{equation*}
Finally, we define the total difference in mean response between the two groups as
\begin{equation}\label{eq:delta}
    \Delta
    = \mu_{Y_A} - \mu_{Y_B}
    = \mathbb{E}(Y_A) - \mathbb{E}(Y_B)
    = \mathbb{E}_{X_A}\{f_A(X_A)\} - \mathbb{E}_{X_B}\{f_B(X_B)\}.
\end{equation}
Two distinct sources contribute to $\Delta$ in Eq.~\eqref{eq:delta}: the groups may differ in the distributions of their covariates ($P_A \neq P_B$) and in their predictor functions ($f_A \neq f_B$). Disentangling these two sources is the object of what follows.

\subsection{Blinder--Oaxaca Decomposition}

The Blinder--Oaxaca decomposition (Blinder, 1973; Oaxaca, 1973) explains the total difference in mean response between the two groups under the linear regression model
\begin{equation}\label{eq:bo-model}
    Y_g = f_{\mathrm{linear},g}(X_g) + \epsilon_g
    = \alpha_g + X_g^\top \beta_g + \epsilon_g,
\end{equation}
where $\alpha_g \in \mathbb{R}$ is the intercept and $\beta_g = (\beta_{g1},\dots,\beta_{gp})^\top \in \mathbb{R}^p$ is the vector of slopes. Under Eq.~\eqref{eq:bo-model}, the marginal mean is given by
\begin{equation*}
    \mu_{Y_g}
    = \mathbb{E}(Y_g)
    = \mathbb{E}_{X_g}\{\mu_{Y_g \mid X_g}\}
    = \alpha_g + \mathbb{E}(X_g)^\top \beta_g
    = \alpha_g + \mu_{X_g}^\top \beta_g.
\end{equation*}
The total difference is then decomposed into structure and composition effects as
\begin{equation}\label{eq:bo-decomp}
\begin{aligned}
    \Delta
    &= \mu_{Y_A}-\mu_{Y_B} \\
    &=
    \underbrace{
        (\alpha_A-\alpha_B)
        + \mu_{X_A}^{\top}(\beta_A-\beta^*)
        + \mu_{X_B}^{\top}(\beta^*-\beta_B)
    }_{\text{$\Lambda^{\mathrm{BO}}$: structure effect}}
    +
    \underbrace{
        (\mu_{X_A}-\mu_{X_B})^{\top}\beta^*
    }_{\text{$\Gamma^{\mathrm{BO}}$: composition effect}} \\
    &=
    \underbrace{
        \Delta\alpha
        + \mu_{X_A}^{\top}(I_p-W)\Delta\beta
        + \mu_{X_B}^{\top}W\Delta\beta
    }_{\text{$\Lambda^{\mathrm{BO}}$: structure effect}}
    +
    \underbrace{
        \Delta\mu_X^{\top} (\beta_B + W\Delta\beta)
    }_{\text{$\Gamma^{\mathrm{BO}}$: composition effect}},
\end{aligned}
\end{equation}
where $\Delta\alpha := \alpha_A-\alpha_B$, $\Delta\mu_X := \mu_{X_A}-\mu_{X_B} = (\Delta\mu_{X_1},\dots,\Delta\mu_{X_p})^\top$, and $\Delta\beta := \beta_A-\beta_B = (\Delta\beta_1,\dots,\Delta\beta_p)^\top$. Here, $\beta^*$ is a reference coefficient vector of the form $\beta^* = W\beta_A + (I_p-W)\beta_B$, where $W = \operatorname{diag}(w_1,\dots,w_p)$ with $0 \leq w_j \leq 1$ for $j = 1,\dots,p$, and $I_p$ denotes the $p \times p$ identity matrix. When $W=0_p$, $I_p$, or $\tfrac{1}{2}I_p$, the reference coefficient vector $\beta^*$ equals $\beta_B$, $\beta_A$, or $\tfrac{1}{2}(\beta_A+\beta_B)$, respectively (Blinder, 1973; Oaxaca, 1973; Reimers, 1983); alternatives include group-share weighting (Cotton, 1988) and pooled-regression coefficients (Neumark, 1988). Note that the structure effect $\Lambda^{\mathrm{BO}}$ reflects differences in the intercept and slopes, $\Delta\alpha$ and $\Delta\beta$, whereas the composition effect $\Gamma^{\mathrm{BO}}$ reflects differences in covariate means, $\Delta\mu_X$, evaluated at $\beta^*$.\\
\indent The decomposition can further be expressed componentwise as
\begin{equation}\label{eq:bo-component}
    \Delta =
    \underbrace{\Delta\alpha}_{\text{$\Lambda_0^{\mathrm{BO}}$: intercept effect}}
    + \sum_{j=1}^p
    \underbrace{
        \bigl\{(1-w_j)\mu_{X_{Aj}} + w_j\mu_{X_{Bj}}\bigr\}\Delta\beta_j
    }_{\text{$\Lambda_j^{\mathrm{BO}}$: structure effect}}
    + \sum_{j=1}^p
    \underbrace{
        \Delta\mu_{X_j}\bigl(\beta_{Bj}+w_j\Delta\beta_j\bigr)
    }_{\text{$\Gamma_j^{\mathrm{BO}}$: composition effect}},
\end{equation}
where $\Lambda_0^{\mathrm{BO}}$ denotes the intercept channel, and $\Lambda_j^{\mathrm{BO}}$ and $\Gamma_j^{\mathrm{BO}}$ denote the structure and composition channels, respectively, for covariate $j$. This componentwise representation provides a complete additive decomposition across channels, $\Delta = \Lambda_0^{\mathrm{BO}} + \sum_{j=1}^p \Lambda_j^{\mathrm{BO}} + \sum_{j=1}^p \Gamma_j^{\mathrm{BO}}$, and satisfies channel-specific vanishing properties: $\Lambda_0^{\mathrm{BO}} = 0$ if $\Delta\alpha = 0$, $\Lambda_j^{\mathrm{BO}} = 0$ if $\Delta\beta_j = 0$, and $\Gamma_j^{\mathrm{BO}} = 0$ if $\Delta\mu_{X_j} = 0$. Each component is also linear in its corresponding difference term, holding the remaining quantities fixed.

\subsection{Game-Theoretic Attribution with Minimum-Switching-Distortion Coupling}

\subsubsection{Cooperative Switching Game}
\label{sec:switching}

Throughout, we take group $B$ as the reference group and group $A$ as the comparison group, so that all switches proceed from group $B$ to group $A$. Exchanging the group labels simply reverses the sign of each effect under an admissible coupling rule, as shown in Section~\ref{sec:properties}.\\
\noindent \textbf{Switches, players, and unions.}
The cooperative switching game is built on two types of switches: a model switch replaces $f_B$ by $f_A$, whereas a covariate switch replaces group-$B$ covariates by their group-$A$ counterparts. A covariate switch is full if all $p$ covariates are switched and partial otherwise.\\
\indent To formalize partial covariate switches, we index the covariates by $[p]:=\{1,\dots,p\}$. For a subset $H\subseteq[p]$ of switched covariates and its complement $H^c:=[p]\setminus H$ of retained covariates, we denote the corresponding subvectors of $X_g=(X_{g1},\dots,X_{gp})^\top$ by
\begin{equation*}
X_{g,H}:=(X_{gj})_{j\in H},
\qquad
X_{g,H^c}:=(X_{gj})_{j\in H^c},
\qquad
g\in\{A,B\},
\end{equation*}
which have dimensions $|H|$ and $p-|H|$, respectively. To encode the source group of each coordinate, we define
\begin{equation*}
\Omega(H)
:=
\{(A,j):j\in H\}
\cup
\{(B,j):j\in H^c\},
\end{equation*}
so that the coordinates in $H$ are taken from group $A$ and those in $H^c$ are retained from group $B$. The corresponding hybrid covariate vector $X_{\Omega(H)}$ is then defined coordinatewise as
\begin{equation}\label{eq:hybrid-vector-def}
\bigl(X_{\Omega(H)}\bigr)_j
:=
\begin{cases}
X_{Aj}, & j\in H,\\[4pt]
X_{Bj}, & j\in H^c,
\end{cases}
\qquad j\in[p].
\end{equation}
Thus, unlike the lower-dimensional subvectors $X_{g,H}$ and $X_{g,H^c}$, $X_{\Omega(H)}$ is a full $p$-dimensional vector that combines $X_{A,H}$ on the switched coordinates with $X_{B,H^c}$ on the retained coordinates, in the original order of the covariates. At the two endpoints, Eq.~\eqref{eq:hybrid-vector-def} gives
\begin{equation}\label{eq:hybrid-vector-endpoints}
X_{\Omega(\emptyset)}=X_B,
\qquad
X_{\Omega([p])}=X_A,
\end{equation}
corresponding to no covariate switch and the full covariate switch, respectively.\\
\indent We then treat the model switch and the $p$ individual covariate switches as players, with
\begin{equation*}
N=\{M,C_1,\dots,C_p\},
\end{equation*}
where $M$ denotes the model switch and $C_j$ the switch of covariate $j$. A coalition $S\subseteq N$ specifies which switches are applied. To apply the Owen value (Owen, 1977), we partition $N$ into two \textit{a priori} unions,
\begin{equation*}
U_M:=\{M\},
\qquad
U_C:=\{C_1,\dots,C_p\},
\qquad
\mathcal{P}:=\{U_M,U_C\},
\end{equation*}
where $U_M$ denotes the singleton model union, $U_C$ denotes the covariate union, and $\mathcal{P}$ denotes the resulting partition.\\
\noindent \textbf{Couplings for hybrid covariate vectors.} At the two endpoints in Eq.~\eqref{eq:hybrid-vector-endpoints}, no coordinates from the two groups are combined: $X_{\Omega(\emptyset)}=X_B \sim P_B$ and $X_{\Omega([p])}=X_A \sim P_A$, so both configurations are fully specified by the endpoint distributions $P_B$ and $P_A$ alone. For every intermediate $H\subseteq[p]$, however, $X_{\Omega(H)}$ combines coordinates of $X_A$ with coordinates of $X_B$, which requires $X_A$ and $X_B$ to be defined on a common probability space. We therefore let $\pi$ denote a coupling of $P_A$ and $P_B$, that is, a joint distribution on $\mathbb{R}^p\times\mathbb{R}^p$ with marginals $P_A$ and $P_B$, respectively, and write
\begin{equation*}
(X_A,X_B)\sim\pi.
\end{equation*}
Thus, $\pi$ specifies the cross-group joint distribution of $X_A$ and $X_B$, and hence the dependence between them. Given $\pi$, the hybrid covariate vector $X_{\Omega(H)}$ is well defined for every $H\subseteq[p]$, and we denote its distribution by
\begin{equation}\label{eq:hybrid-distribution}
Q_{\Omega(H)}^\pi
:=
\mathcal{L}_{\pi}\bigl(X_{\Omega(H)}\bigr),
\qquad
H\subseteq[p].
\end{equation}
The $p$ covariate switches yield $2^p$ hybrid covariate vectors. The coupling $\pi$ is held fixed across coalitions, while the induced hybrid covariate distribution varies with $H$. The rule used to select $\pi$ is specified in Section~\ref{sec:coupling}.\\
\noindent \textbf{Characteristic function.} For any coalition $S\subseteq N$, we define
\begin{equation*}
H(S):=\{j\in[p]:C_j\in S\},
\qquad
g(S):=
\begin{cases}
A, & M\in S,\\
B, & M\notin S,
\end{cases}
\end{equation*}
so that $H(S)$ determines which covariates are switched, whereas $g(S)$ determines which predictor function is used. Given $\pi$, we define the characteristic function $v^\pi:2^N\to\mathbb{R}$ by
\begin{equation}\label{eq:full-game}
v^\pi(S)
:=
\mathbb{E}_{\pi}
\bigl\{
f_{g(S)}(X_{\Omega(H(S))})
\bigr\}
-
\mu_{Y_B},
\qquad S\subseteq N.
\end{equation}
Thus, each coalition value is the change in the mean response relative to the baseline configuration of group $B$. At the two endpoints, Eqs.~\eqref{eq:hybrid-vector-endpoints} and \eqref{eq:full-game} give
\begin{equation}\label{eq:empty-grand-values}
v^\pi(\emptyset)
=
\mathbb{E}\{f_B(X_B)\}-\mu_{Y_B}
=
0,
\qquad
v^\pi(N)
=
\mathbb{E}\{f_A(X_A)\}-\mu_{Y_B}
=
\mu_{Y_A}-\mu_{Y_B}
=
\Delta
\end{equation}
for any coupling $\pi$, corresponding to the empty coalition, which applies no switch, and the grand coalition, which applies all switches and returns the total difference $\Delta$ in Eq.~\eqref{eq:delta}, respectively.

\subsubsection{Two-Stage Hierarchical Allocation via the Owen Value}
\label{sec:game}
 
In our cooperative switching game $(N,v^\pi)$, the Owen value (Owen, 1977) first allocates $\Delta$ between the unions $U_M$ and $U_C$ via the quotient game and then allocates the payoff of $U_C$ among $C_1,\dots,C_p$ via the internal game.\\
\noindent \textbf{Stage 1: quotient game between the unions.} In Stage~1, $U_M$ and $U_C$ are the two players of the quotient game associated with $\mathcal{P}$ (Owen, 1977), and the payoff of each union is its average marginal contribution over the two possible orders of the unions (Shapley, 1953). The marginal contributions involve only the four coalitions $\emptyset$, $U_M$, $U_C$, and $N$, in which either no covariate or all covariates are switched. Along with $v^\pi(\emptyset)=0$ and $v^\pi(N)=\Delta$ in Eq.~\eqref{eq:empty-grand-values}, Eqs.~\eqref{eq:hybrid-vector-endpoints} and \eqref{eq:full-game} give
\begin{equation}\label{eq:endpoint-values}
v^\pi(U_M)
=
\mathbb{E}\{f_A(X_B)\}-\mu_{Y_B},
\qquad
v^\pi(U_C)
=
\mathbb{E}\{f_B(X_A)\}-\mu_{Y_B},
\end{equation}
where $v^\pi(U_M)$ switches the predictor function alone, whereas $v^\pi(U_C)$ switches the covariate vector alone. We define the aggregate structure and composition effects as the payoffs of $U_M$ and $U_C$, respectively, which, by Eqs.~\eqref{eq:empty-grand-values} and \eqref{eq:endpoint-values}, are
\begin{equation}\label{eq:aggregate-effects}
\begin{aligned}
\Lambda
&:=
\frac{1}{2}
\left[
v^\pi(U_M)-v^\pi(\emptyset)
+
v^\pi(N)-v^\pi(U_C)
\right]
=
\frac{1}{2}\mathbb{E}_{P_A}\{d(X)\}
+
\frac{1}{2}\mathbb{E}_{P_B}\{d(X)\},\\[4pt]
\Gamma
&:=
\frac{1}{2}
\left[
v^\pi(U_C)-v^\pi(\emptyset)
+
v^\pi(N)-v^\pi(U_M)
\right]
=
\mathbb{E}_{P_A}\{\bar f(X)\}
-
\mathbb{E}_{P_B}\{\bar f(X)\},
\end{aligned}
\end{equation}
where $d(x):=f_A(x)-f_B(x)$ and $\bar f(x):=\{f_A(x)+f_B(x)\}/2$ denote the difference and midpoint of the two predictor functions, respectively. The structure effect $\Lambda$ averages the change from $f_B$ to $f_A$ under the two endpoint covariate distributions, whereas the composition effect $\Gamma$ averages the change from $X_B$ to $X_A$ under the two group-specific predictor functions. Because both effects depend only on the endpoint predictor functions and covariate distributions, Stage~1 is coupling-invariant. By efficiency of the Shapley value (Shapley, 1953), the two union payoffs exhaust the total difference,
\begin{equation}\label{eq:aggregate-efficiency}
\Delta
=
\Lambda+\Gamma,
\end{equation}
which is the two-factor Shapley decomposition of $\Delta$ in the sense of Shorrocks (2013).\\
\noindent \textbf{Stage 2: internal game within the covariate union.} In Stage~2, $C_1,\dots,C_p$ are the players of the internal game associated with $U_C$, and the payoff of each covariate-switch player is its average marginal contribution over all possible orders of the covariate-switch players and the two possible orders of the unions (Owen, 1977). The marginal contributions involve the coalitions $\mathcal{C}_H:=\{C_j:j\in H\}$ and $U_M\cup\mathcal{C}_H$ for $H\subseteq[p]$. Given $\pi$, we define the covariate-specific composition effect of covariate $j$ as the payoff of $C_j$, which, by Eq.~\eqref{eq:full-game}, is
\begin{equation}\label{eq:gamma-j}
\begin{aligned}
\Gamma_j^\pi
&:=
\sum_{H\subseteq[p]\setminus\{j\}}
\frac{|H|!\,(p-|H|-1)!}{p!}
\frac{1}{2}
\Bigl[
v^\pi(\mathcal{C}_{H\cup\{j\}})-v^\pi(\mathcal{C}_H)\\
&\hspace{52mm}
+
v^\pi(U_M\cup\mathcal{C}_{H\cup\{j\}})
-
v^\pi(U_M\cup\mathcal{C}_H)
\Bigr]\\
&=
\sum_{H\subseteq[p]\setminus\{j\}}
\frac{|H|!\,(p-|H|-1)!}{p!}
\left[
\mathbb{E}_{\pi}
\bigl\{
\bar f(X_{\Omega(H\cup\{j\})})
\bigr\}
-
\mathbb{E}_{\pi}
\bigl\{
\bar f(X_{\Omega(H)})
\bigr\}
\right],
\qquad j\in[p].
\end{aligned}
\end{equation}
The second equality follows by averaging the two marginal contributions under $f_B$ and $f_A$, yielding $\bar f$. Thus, while $\Gamma$ measures the effect of switching all covariates, $\Gamma_j^\pi$ is the Shapley-weighted average marginal effect of switching covariate $j$ over $H\subseteq[p]\setminus\{j\}$. Because $\Gamma_j^\pi$ depends on the intermediate hybrid distributions induced by $\pi$ in Eq.~\eqref{eq:hybrid-distribution}, Stage~2 is generally coupling-dependent. By the quotient-game property of the Owen value (Owen, 1977), for any coupling $\pi$, the payoffs of $C_1,\dots,C_p$ exhaust the payoff of $U_C$,
\begin{equation*}
\sum_{j=1}^p\Gamma_j^\pi=\Gamma.
\end{equation*}
Thus, Stage~2 refines $\Gamma$ into covariate-specific components without changing the Stage~1 split.\\
\indent The construction above identifies $\Lambda$ and $\Gamma_1^\pi,\dots,\Gamma_p^\pi$ as the Owen payoffs of the model-switch and covariate-switch players, respectively. Beyond the two-level identities established above, the allocation inherits axioms for fair allocation from the Owen value, including null-player, within-union symmetry, and linearity (Owen, 1977). These results are summarized formally in the following proposition.
 
\begin{proposition}[Owen representation and its consequences]
\label{prop:owen-representation}
For any coupling $\pi$, the Owen allocation of $(N,v^\pi)$ under $\mathcal{P}$ assigns $\Lambda$ to $M$ and $\Gamma_j^\pi$ to $C_j$, for $j\in[p]$, and satisfies:
\begin{enumerate}[label=(\roman*)]
\item \emph{Two-level coherence.} The covariate-specific composition effects exhaust the covariate-union payoff: $\sum_{j=1}^p\Gamma_j^\pi=\Gamma$.
\item \emph{Efficiency.} The structure effect and covariate-specific composition effects together exhaust the total difference: $\Delta=\Lambda+\sum_{j=1}^p\Gamma_j^\pi$.
\item \emph{Null player.} If $v^\pi(S\cup\{C_j\})=v^\pi(S)$ for every $S\subseteq N\setminus\{C_j\}$, then $\Gamma_j^\pi=0$.
\item \emph{Symmetry within the covariate union.} If $v^\pi(S\cup\{C_j\})=v^\pi(S\cup\{C_k\})$ for every $S\subseteq N\setminus\{C_j,C_k\}$, then $\Gamma_j^\pi=\Gamma_k^\pi$.
\item \emph{Linearity.} If $v^\pi=a_1v'+a_2v''$, then each payoff is the corresponding linear combination of the Owen payoffs under $v'$ and $v''$ with the same union structure $\mathcal{P}$.
\end{enumerate}
The proof is given in \SM{S1}.
\end{proposition}

\indent The \textit{a priori} union structure preserves the aggregate interpretation. Under a flat Shapley allocation on the full player set, intermediate hybrid covariate vectors can enter the payoff of $M$, so the model-switch payoff need not equal $\Lambda$ and the covariate-switch payoffs need not sum to $\Gamma$. The hierarchical Owen allocation avoids this by preserving the quotient-game split and restricting the covariate-specific refinement to $U_C$. A formal comparison is given in \SM{S2}.\\
\indent Because Proposition~\ref{prop:owen-representation} holds for any coupling $\pi$, only the internal-game allocation depends on the choice of $\pi$ through the intermediate hybrid covariate vectors. We specify this choice next.
 
\subsubsection{Admissible Coupling Rules and the Minimum-Switching-Distortion Coupling}
\label{sec:coupling}
 
\noindent \textbf{Indeterminacy of the hybrid covariate distribution.} By Eq.~\eqref{eq:hybrid-vector-def}, $X_{\Omega(H)}$ combines the switched block $X_{A,H}$ with the retained block $X_{B,H^c}$, whose distributions are denoted by $P_{A,H}$ and $P_{B,H^c}$, respectively. By the marginal constraints of the coupling, the hybrid covariate distribution $Q_{\Omega(H)}^\pi$ in Eq.~\eqref{eq:hybrid-distribution} satisfies the block-marginal constraints
\begin{equation*}
\mathcal{L}_{\pi}\bigl((X_{\Omega(H)})_H\bigr)=P_{A,H},
\qquad
\mathcal{L}_{\pi}\bigl((X_{\Omega(H)})_{H^c}\bigr)=P_{B,H^c}.
\end{equation*}
Thus, $P_A$ and $P_B$ determine the dependence within each block but not the cross-block dependence, because they do not specify how $X_A$ is paired with $X_B$. For an intermediate $H$, $Q_{\Omega(H)}^\pi$ is therefore generally not determined by $P_A$ and $P_B$ alone, and this cross-block dependence can affect $\Gamma_j^\pi$ in Eq.~\eqref{eq:gamma-j} when $\bar f$ is nonadditive. We use a single coupling $\pi$ to specify this cross-block dependence across all $H$, rather than a separate hybrid distribution for each $H$. Holding $\pi$ fixed across coalitions ensures that each marginal contribution in Eq.~\eqref{eq:gamma-j} reflects only a covariate switch, not a simultaneous change in cross-group dependence. It remains to specify how $\pi$ is selected.\\
\noindent \textbf{Admissible coupling rules.} Not every coupling rule is equally suitable for the decomposition. We call a coupling rule admissible if it is consistent at coincident endpoint distributions and equivariant to group exchange and common covariate relabeling, as formalized below.

\begin{definition}[Admissible coupling rule]\label{def:admissible-coupling}
A coupling rule on a domain $\mathcal{D}$ maps each $(P_A,P_B)\in\mathcal{D}$ to a coupling of $P_A$ and $P_B$. We call the rule admissible if it satisfies the following conditions.
\begin{enumerate}[label=(\roman*)]
\item \emph{Identity at coincidence.} If $(P,P)\in\mathcal{D}$, the rule selects the identity coupling $\mathcal{L}(X,X)$ with $X\sim P$, so that $X_A=X_B$ almost surely.
\item \emph{Transposition equivariance.} If $(P_A,P_B)\in\mathcal{D}$ and $(P_B,P_A)\in\mathcal{D}$, exchanging $P_A$ and $P_B$ transposes the selected coupling.
\item \emph{Coordinate equivariance.} If a common relabeling of the covariates maps $(P_A,P_B)\in\mathcal{D}$ to another pair in $\mathcal{D}$, the selected coupling is relabeled in the same way.
\end{enumerate}
\end{definition}
 
\indent Condition (i) selects the identity coupling when $P_A=P_B$, so $X_{\Omega(H)}=X_A=X_B$ almost surely for every $H\subseteq[p]$ and the internal-game coalition value is invariant to the covariate-switch subset. Conditions (ii) and (iii) ensure consistency under group exchange and common covariate relabeling, respectively. These conditions restrict the coupling rule without necessarily identifying a unique coupling.\\
\noindent \textbf{MSD coupling.} We select $\pi$ by the MSD criterion, formulated as an optimal-transport problem that minimizes the expected squared standardized distance between paired $X_A$ and $X_B$. Specifically, we compare couplings by the expected switching cost
\begin{equation}\label{eq:switching-cost}
\mathbb{E}_\pi\{c(X_A,X_B)\},
\end{equation}
where
\begin{equation*}
c(x_A,x_B)
:=
\sum_{j\in J}
\left(
\frac{x_{Aj}-x_{Bj}}{s_j}
\right)^2,
\qquad
x_A,x_B\in\mathbb{R}^p,
\end{equation*}
with
\begin{equation*}
s_j^2
:=
\frac{1}{2}
\left\{
(\Sigma_{X_A})_{jj}
+
(\Sigma_{X_B})_{jj}
\right\},
\qquad
J:=\{j\in[p]:s_j>0\}.
\end{equation*}
Here, $s_j$ is a group-symmetric scale, and $J$ indexes covariates that are nonconstant in at least one group. For $j\notin J$, the covariate is constant within each group, so its paired difference is coupling-invariant; it is therefore excluded from the cost but remains a covariate-switch player. The symmetric scaling and coordinatewise sum make the criterion invariant to group exchange and common covariate relabeling. Standardization also makes it invariant to common coordinatewise affine transformations $x_j\mapsto a_j+b_jx_j$ with $b_j\neq0$. The criterion depends only on the covariate distributions, not on the response or predictor functions.
 
\begin{definition}[MSD coupling]\label{def:msd-coupling}
We define the MSD coupling $\pi^\star$ as a minimizer of Eq.~\eqref{eq:switching-cost} over all couplings of $P_A$ and $P_B$,
\begin{equation}\label{eq:msd-coupling}
\pi^\star
\in
\operatorname*{arg\,min}_{\pi}
\mathbb{E}_\pi\{c(X_A,X_B)\}.
\end{equation}
If multiple minimizers exist, $\pi^\star$ is selected by a prespecified deterministic tie-breaking rule; the empirical implementation uses the maximum-entropy rule in Section~\ref{sec:computation}.
\end{definition}
 
\indent After standardizing the coordinates in $J$ by $x_j\mapsto x_j/s_j$, Eq.~\eqref{eq:msd-coupling} is the Kantorovich optimal-transport problem on $\mathbb{R}^{|J|}$ with squared Euclidean cost (Villani, 2009; Peyr\'e and Cuturi, 2019). Finite second moments ensure existence, while absolute continuity of either standardized endpoint distribution is a sufficient condition for uniqueness (Brenier, 1991; McCann, 1995). Details are given in \SM{S3}.\\
\indent The invariances above yield conditions (ii) and (iii) of Definition~\ref{def:admissible-coupling}. When $P_A=P_B$, the identity coupling uniquely attains zero switching cost, yielding condition (i). Hence:
 
\begin{proposition}[Admissibility]\label{prop:msd-admissible}
On any domain where the MSD minimizer in Eq.~\eqref{eq:msd-coupling} is unique, the coupling rule $(P_A,P_B)\mapsto\pi^\star$ is admissible. Under nonuniqueness, admissibility holds if the tie-breaking rule is equivariant to group exchange and common covariate relabeling. The proof is given in \SM{S3}.
\end{proposition}
 
\indent MAD uses the MSD coupling $\pi^\star$, with covariate-specific composition effects
\begin{equation}\label{eq:MAD-gamma-j}
\Gamma_j
:=
\Gamma_j^{\pi^\star},
\qquad
j\in[p],
\end{equation}
so that $\Delta=\Lambda+\sum_{j=1}^p\Gamma_j$ by Proposition~\ref{prop:owen-representation}.\\
\noindent \textbf{Comparison with independence and entropically regularized couplings.}
The independence coupling $\pi^{\mathrm{ind}}:=P_A\otimes P_B$ pairs $X_A$ and $X_B$ independently, yielding $Q_{\Omega(H)}^{\pi^{\mathrm{ind}}}=P_{A,H}\otimes P_{B,H^c}$ and thereby removing dependence between the switched and retained blocks. For $\varepsilon>0$, the entropically regularized coupling is $\pi^\varepsilon:=\operatorname*{arg\,min}_{\pi}\bigl[\mathbb{E}_\pi\{c(X_A,X_B)\}+\varepsilon\,\mathrm{KL}(\pi\,\|\,P_A\otimes P_B)\bigr]$, where $\mathrm{KL}$ denotes the Kullback--Leibler divergence. It approaches $\pi^{\mathrm{ind}}$ as $\varepsilon\to\infty$ and, when the MSD minimizer is unique, $\pi^\star$ as $\varepsilon\to0$ (Cominetti and San Mart\'in, 1994; Cuturi, 2013; Peyr\'e and Cuturi, 2019). Both alternatives satisfy conditions (ii) and (iii) of Definition~\ref{def:admissible-coupling} but violate condition (i) whenever the common distribution is not a point mass (\SM{S4}). Thus, when $P_A=P_B$, transposition equivariance still gives $\Gamma_j^\pi=0$ for every $j\in[p]$, but partial covariate switches can change the internal-game coalition values when $\bar f$ is nonadditive (Section~\ref{sec:properties}).
 
\subsubsection{Further Properties and Connection to the Blinder--Oaxaca Decomposition}
\label{sec:properties}
 
We establish further properties of MAD concerning coupling dependence, null cases, and group exchange.
 
\begin{proposition}[Further properties of MAD]\label{prop:basic-properties}
MAD satisfies the following properties.
\begin{enumerate}[label=(\roman*)]
\item \emph{Coupling invariance of the aggregate effects.} The aggregate effects $\Lambda$ and $\Gamma$ do not depend on the coupling.
\item \emph{Coupling invariance under an additive midpoint predictor.} If $\bar f(x)=c_0+\sum_{j=1}^p\psi_j(x_j)$, where $c_0$ is constant and the univariate functions $\psi_1,\dots,\psi_p$ have finite expectations under $P_A$ and $P_B$, then $\Gamma_j^\pi=\mathbb{E}_{P_A}\{\psi_j(X_j)\}-\mathbb{E}_{P_B}\{\psi_j(X_j)\}$ for any coupling $\pi$ and $j\in[p]$.
\item \emph{Null structure effect.} If $f_A(x)=f_B(x)$ for $P_A$- and $P_B$-almost every $x$, then $\Lambda=0$.
\item \emph{Null composition effects at coincident distributions.} If $P_A=P_B=P$, then $\Gamma=0$ for any coupling $\pi$. If the coupling is selected by a rule satisfying condition (ii) of Definition~\ref{def:admissible-coupling}, then $\Gamma_j^\pi=0$ for every $j\in[p]$. If the rule satisfies condition (i), then $\mathbb{E}_\pi\{\bar f(X_{\Omega(H)})\}=\mathbb{E}_P\{\bar f(X)\}$ for every $H\subseteq[p]$.
\item \emph{Group-exchange equivariance.} If the coupling is selected by a rule satisfying condition (ii) of Definition~\ref{def:admissible-coupling}, exchanging groups $A$ and $B$ reverses the signs of $\Lambda$, $\Gamma$, and each covariate-specific composition effect.
\end{enumerate}
The proof is given in \SM{S5}.
\end{proposition}
 
\indent Parts (i) and (ii) characterize the role of the coupling: it does not affect $\Lambda$ or $\Gamma$, and can affect the covariate-specific composition effects only when $\bar f$ is nonadditive. Part (iv) separates two roles at coincidence: transposition equivariance makes the covariate-specific composition effects vanish, whereas identity at coincidence also makes the internal-game coalition value invariant to the covariate-switch subset. The MSD coupling satisfies both conditions by Proposition~\ref{prop:msd-admissible}.\\
\indent We next show that, under linear predictor functions, MAD reduces to the Blinder--Oaxaca decomposition with the midpoint reference coefficient vector of Reimers (1983).
 
\begin{proposition}[Reduction to the Blinder--Oaxaca decomposition]\label{prop:bo}
Suppose that $f_g(x)=\alpha_g+x^\top\beta_g$ for $g\in\{A,B\}$, and let $\Delta\alpha$, $\Delta\beta_j$, and $\Delta\mu_{X_j}$ be as in Eq.~\eqref{eq:bo-decomp}. Then, for any coupling $\pi$,
\begin{equation*}
\Lambda
=
\Delta\alpha
+
\sum_{j=1}^p
\frac{\mu_{X_{Aj}}+\mu_{X_{Bj}}}{2}\Delta\beta_j,
\qquad
\Gamma_j^\pi
=
\Delta\mu_{X_j}\frac{\beta_{Aj}+\beta_{Bj}}{2},
\qquad j\in[p].
\end{equation*}
Hence, with the components in Eq.~\eqref{eq:bo-component} evaluated at $W=\tfrac12 I_p$, $\Lambda=\Lambda_0^{\mathrm{BO}}+\sum_{j=1}^p\Lambda_j^{\mathrm{BO}}$ and $\Gamma_j^\pi=\Gamma_j^{\mathrm{BO}}$ for every $j\in[p]$. The proof is given in \SM{S6}.
\end{proposition}

\indent The reduction holds for any coupling $\pi$ because linear predictor functions yield an additive midpoint predictor, so Proposition~\ref{prop:basic-properties}(ii) applies. Thus, MAD retains the classical decomposition as a special case, with the midpoint reference arising from the equal weighting of the two union orders, in line with the Shapley decomposition of Shorrocks (2013).

\subsubsection{Computational Procedures}
\label{sec:computation}
 
For each group $g\in\{A,B\}$, we let $\{(Y_{gi},X_{gi}):i=1,\dots,n_g\}$ be an independent random sample, and let $\hat f_A$ and $\hat f_B$ denote the fitted predictor functions. Treating the fitted predictor functions as given, we compute the plug-in cross-evaluated means as
\begin{equation*}
\hat\mu_{r\mid s}
:=
\frac{1}{n_s}
\sum_{i=1}^{n_s}
\hat f_r(X_{si}),
\qquad
r,s\in\{A,B\}.
\end{equation*}
The total difference and aggregate effects are then
\begin{equation*}
\hat\Delta
:=
\hat\mu_{A\mid A}-\hat\mu_{B\mid B},
\qquad
\hat\Lambda
:=
\tfrac12\bigl(\hat\mu_{A\mid A}+\hat\mu_{A\mid B}-\hat\mu_{B\mid A}-\hat\mu_{B\mid B}\bigr),
\qquad
\hat\Gamma
:=
\tfrac12\bigl(\hat\mu_{A\mid A}+\hat\mu_{B\mid A}-\hat\mu_{A\mid B}-\hat\mu_{B\mid B}\bigr).
\end{equation*}
Thus, $\hat\Delta=\hat\Lambda+\hat\Gamma$ exactly, and these aggregate quantities require no coupling.\\
\noindent \textbf{Empirical MSD coupling.}
We let $\widehat P_g$ denote the empirical covariate distribution of group $g$, which assigns mass $1/n_g$ to each $X_{gi}$, and define
\begin{equation*}
\hat s_j^2
:=
\tfrac12\bigl(\hat\sigma_{Aj}^2+\hat\sigma_{Bj}^2\bigr),
\qquad
\hat J:=\{j\in[p]:\hat s_j>0\},
\end{equation*}
where $\hat\sigma_{gj}^2$ is the empirical variance computed with divisor $n_g$. The pairwise switching costs are
\begin{equation}\label{eq:empirical-switching-cost}
C_{ii'}
:=
\sum_{j\in\hat J}
\left(
\frac{X_{Aij}-X_{Bi'j}}{\hat s_j}
\right)^2,
\qquad
i=1,\dots,n_A,\quad
i'=1,\dots,n_B.
\end{equation}
We let $\mathcal{U}_{n_A,n_B}$ be the set of nonnegative $n_A\times n_B$ matrices $\hat\pi=(\hat\pi_{ii'})$ with row sums $1/n_A$ and column sums $1/n_B$, representing couplings of $\widehat P_A$ and $\widehat P_B$. The empirical MSD problem with ground cost $C_{ii'}$ is
\begin{equation}\label{eq:empirical-msd}
\widehat{\Pi}_{\mathrm{MSD}}
:=
\operatorname*{arg\,min}_{\hat\pi\in\mathcal{U}_{n_A,n_B}}
\sum_{i=1}^{n_A}
\sum_{i'=1}^{n_B}
\hat\pi_{ii'}C_{ii'}.
\end{equation}
We select the maximum-entropy element of $\widehat{\Pi}_{\mathrm{MSD}}$, which is obtained as the $\varepsilon\to0$ limit of the corresponding entropically regularized transport problem (Cominetti and San Mart\'in, 1994):
\begin{equation*}
\hat\pi^\star
:=
\operatorname*{arg\,max}_{\hat\pi\in\widehat{\Pi}_{\mathrm{MSD}}}
\mathcal{H}(\hat\pi),
\qquad
\mathcal{H}(\hat\pi)
:=
-\sum_{i=1}^{n_A}
\sum_{i'=1}^{n_B}
\hat\pi_{ii'}\log\hat\pi_{ii'},
\end{equation*}
with $0\log0:=0$. Strict concavity of $\mathcal{H}$ on the convex set $\widehat{\Pi}_{\mathrm{MSD}}$ ensures uniqueness, and the resulting tie-breaking rule is equivariant as required by Proposition~\ref{prop:msd-admissible} (\SM{S3}).\\
\noindent \textbf{Empirical independence and entropically regularized couplings.}
The empirical independence coupling $\hat\pi^{\mathrm{ind}}:=\widehat P_A\otimes\widehat P_B$ assigns $\hat\pi^{\mathrm{ind}}_{ii'}=1/(n_An_B)$ to every pair $(i,i')$ and requires no transport optimization or cost matrix. For $\varepsilon>0$, the empirical entropically regularized coupling is $\hat\pi^{\varepsilon}:=\operatorname*{arg\,min}_{\hat\pi\in\mathcal{U}_{n_A,n_B}}\bigl\{\sum_{i,i'}\hat\pi_{ii'}C_{ii'}-\varepsilon\mathcal{H}(\hat\pi)\bigr\}$, the empirical counterpart of $\pi^\varepsilon$, since $-\mathcal{H}(\hat\pi)$ differs from $\mathrm{KL}(\hat\pi\,\|\,\hat\pi^{\mathrm{ind}})$ only by the constant $\log(n_An_B)$ on $\mathcal{U}_{n_A,n_B}$. The coupling $\hat\pi^{\varepsilon}$ is unique and can be computed by Sinkhorn iterations (Cuturi, 2013; Peyr\'e and Cuturi, 2019). Both couplings have full support on all $n_An_B$ pairs.\\
\noindent \textbf{Exact and approximate allocations.}
For each $H\subseteq[p]$ and observation pair $(i,i')$, we define
\begin{equation*}
\bigl(X_{\Omega(H)}(i,i')\bigr)_j
:=
\begin{cases}
X_{Aij}, & j\in H,\\
X_{Bi'j}, & j\in H^c,
\end{cases}
\qquad
j\in[p].
\end{equation*}
We let $\hat{\bar f}:=(\hat f_A+\hat f_B)/2$ be the fitted midpoint predictor function and define
\begin{equation*}
\hat m_{\hat\pi^\star}(H)
:=
\sum_{i=1}^{n_A}
\sum_{i'=1}^{n_B}
\hat\pi^\star_{ii'}
\hat{\bar f}\!\left(X_{\Omega(H)}(i,i')\right),
\qquad
H\subseteq[p].
\end{equation*}
Since $\hat\pi^\star\in\mathcal{U}_{n_A,n_B}$, $\hat m_{\hat\pi^\star}([p])-\hat m_{\hat\pi^\star}(\emptyset)=\hat\Gamma$. The exact covariate-specific allocation is
\begin{equation*}
\hat\Gamma_j
:=
\sum_{H\subseteq[p]\setminus\{j\}}
\frac{|H|!\,(p-|H|-1)!}{p!}
\left[
\hat m_{\hat\pi^\star}(H\cup\{j\})
-
\hat m_{\hat\pi^\star}(H)
\right],
\qquad
j\in[p].
\end{equation*}
\indent For larger $p$, the approximate allocation averages marginal contributions over sampled permutations (Castro et al., 2009; Mitchell et al., 2022). We let $\rho_1,\dots,\rho_R$ be independent uniform permutations of $[p]$ and $\operatorname{Pre}_j(\rho_m)$ the set of covariates preceding $j$ in $\rho_m$, and we compute
\begin{equation}\label{eq:composition-shapley-mc}
\hat\Gamma_j^{(R)}
:=
\frac{1}{R}
\sum_{m=1}^R
\left[
\hat m_{\hat\pi^\star}\!\left(\operatorname{Pre}_j(\rho_m)\cup\{j\}\right)
-
\hat m_{\hat\pi^\star}\!\left(\operatorname{Pre}_j(\rho_m)\right)
\right],
\qquad
j\in[p].
\end{equation}
Since all covariates share the same permutations, the marginal contributions telescope within each permutation, so that $\sum_{j=1}^p\hat\Gamma_j^{(R)}=\hat\Gamma$ and $\hat\Delta=\hat\Lambda+\sum_{j=1}^p\hat\Gamma_j^{(R)}$ for every finite $R$. The budget $R$ may be fixed or adaptive: pilot permutations determine a bounded budget targeting a prespecified relative Monte Carlo precision, and the allocation is computed from new independent permutations. For a fixed budget, and conditional on the pilot for an adaptive one, Eq.~\eqref{eq:composition-shapley-mc} is unbiased for the exact allocation (\SM{S7}). For $\hat\pi^{\mathrm{ind}}$ and $\hat\pi^{\varepsilon}$, $\hat m_{\hat\pi}(H)$ and the allocations are defined analogously, with $\hat\pi$ in place of $\hat\pi^\star$.\\
\noindent \textbf{Computational complexity.}
For computational comparison, suppose $n_A=n_B=n$. Then Eq.~\eqref{eq:empirical-msd} reduces to a linear assignment problem (Kuhn, 1955), and under a unique optimal assignment the empirical MSD coupling has support size $n$. In contrast, the independence and entropically regularized couplings have support size $n^2$, with the independence coupling requiring no transport optimization or cost-matrix construction. Table~\ref{tab:computational-complexity} summarizes the resulting leading costs. The smaller support of the empirical MSD coupling reduces the allocation cost by a factor of $n$ relative to these alternatives, which lowers the overall cost whenever allocation dominates coupling construction. Additional details are given in \SM{S7}.

\section{Results}
\label{sec:results}

\subsection{Simulation Designs}
\label{sec:simulation-design}
 
We conducted two simulation studies. First, the decomposition study evaluated MAD across predictive models and covariate--response structures, focusing on the recovery of population structure and composition effects and computational cost. Second, the coupling study compared the MSD coupling with the independence and entropically regularized couplings when the covariate distributions coincide, focusing on identity at coincidence, coupling sparsity, and computational cost.\\
\noindent \textbf{Covariate dimensions and sample sizes.}
We considered $p\in\{5,10,15,20\}$ in both the decomposition and coupling studies, with $n_A=n_B=n\in\{100,200\}$ and an independent training sample of size $n_{\mathrm{train}}=1000$ per group. The corresponding coupling and allocation configurations are summarized in Table~\ref{tab:sim-MAD-configurations}.\\
\noindent \textbf{Covariate distributions and predictor functions.}
We generated the covariate vectors as
\begin{equation}\label{eq:sim-covariate-distributions}
X_A\sim N_p(\delta,\Sigma_{X_A}),
\qquad
X_B\sim N_p(0,\Sigma_{X_B}),
\end{equation}
where $\delta$ is the between-group mean shift and $\Sigma_{X_A}$ and $\Sigma_{X_B}$ are the covariance matrices. We also considered four covariate--response structures: linear additive (LA), linear multiplicative (LM), nonlinear additive (NA), and nonlinear multiplicative (NM). Their true predictor functions are summarized in Table~\ref{tab:sim-response-structures}. The coefficient and distributional parameters in Eq.~\eqref{eq:sim-covariate-distributions} and Table~\ref{tab:sim-response-structures} are specified below.\\
\indent In the decomposition study, we set
\begin{equation*}
\beta=\frac{\sqrt{p}}{2}\,\nu_1,
\qquad
\gamma=\frac{\sqrt{p}}{2}\,\nu_2,
\qquad
\delta=\frac{\sqrt{p}}{2}\,(\nu_1+\nu_2+\nu_3),
\qquad
\alpha=1,
\end{equation*}
where $\nu_1,\nu_2,\nu_3$ were obtained by orthonormalizing three independent $N_p(0,I_p)$ draws. Thus $\|\beta\|_2=\|\gamma\|_2=\sqrt{p}/2$, $\|\delta\|_2=\sqrt{3p}/2$, $\beta^\top\gamma=0$, and $\beta^\top\delta=\gamma^\top\delta=p/4$. We let $K=2p/5$, generated $\widetilde\theta$ and $\widetilde\xi$ independently as normalized $N_K(0,I_K)$ vectors, and set $\theta=\sqrt{2p}\,\widetilde\theta$ and $\xi=\sqrt{2p}\,\widetilde\xi$, so that $\|\theta\|_2=\|\xi\|_2=\sqrt{2p}$. Finally, we specified $\Sigma_{X_A}$ and $\Sigma_{X_B}$ as independently generated unstructured correlation matrices by setting $G_g=I_p+(2/q_p)\sum_{k=1}^{q_p}z_{gk}z_{gk}^{\top}$ and $\Sigma_{X_g}=D_g^{-1/2}G_gD_g^{-1/2}$, where $D_g=\operatorname{diag}(G_g)$, $z_{gk}\stackrel{\mathrm{iid}}{\sim}N_p(0,I_p)$, and $q_p=\lceil p/2\rceil$, for $g\in\{A,B\}$.\\ 
\indent In the coupling study, we set $\delta=0$ and $\Sigma_{X_A}=\Sigma_{X_B}=\Sigma_X$, where $\Sigma_X$ was generated using the same unstructured correlation construction, while retaining all other settings from the decomposition study, so that $P_A=P_B=P=N_p(0,\Sigma_X)$, as in Proposition~\ref{prop:basic-properties}(iv). The same covariate sample was used for both groups with $X_A=X_B$.\\
\indent In both studies, the coefficient directions and unstructured correlation matrices were regenerated in each simulation replicate. In the decomposition study, the resulting $\delta$ was also replicate-specific, whereas the coupling study fixed $\delta=0$. Within each replicate, the realized coefficients and correlation matrices were shared across the two sample sizes and all covariate--response structures and, in the coupling study, across all coupling methods.\\
\noindent \textbf{Response generation.}
For $g\in\{A,B\}$ and $\ell\in\{\mathrm{LA},\mathrm{LM},\mathrm{NA},\mathrm{NM}\}$, we generated
\begin{equation*}
Y_g=\eta_g^{(\ell)}(X_g)+\epsilon_g,
\qquad
\epsilon_g\sim N(0,1).
\end{equation*}
\noindent \textbf{Predictive models.}
We considered five predictive models: linear regression; gradient boosting machines (Friedman, 2001), implemented using \texttt{XGBoost} (Chen and Guestrin, 2016) and \texttt{LightGBM} (Ke et al., 2017); and deep neural networks (LeCun et al., 2015), implemented using \texttt{Keras} (Chollet et al., 2015) and \texttt{Torch} (Paszke et al., 2019). Detailed training specifications are provided in \SM{S8}.\\
\indent In addition, we considered an oracle benchmark using the true group-specific predictor functions, $f_g=\eta_g^{(\ell)}$. It removes model-estimation error, so the remaining error arises only from empirical covariate distributions and, under approximate allocation, permutation sampling. The coupling study used the oracle benchmark only.\\
\noindent \textbf{Evaluation criteria.}
The evaluation criteria are summarized in Table~\ref{tab:sim-evaluation}. Each configuration was replicated $B=100$ times.\\
\indent In the decomposition study, we measured the recovery of the aggregate and covariate-specific effects using the relative root mean squared errors (RRMSEs)
\begin{gather*}
\operatorname{RRMSE}_{\mathrm{agg}}
=
\left[
\frac{
\sum_{b=1}^{B}
\left\{
(\hat\Lambda^{(b)}-\Lambda^{0,(b)})^2
+
(\hat\Gamma^{(b)}-\Gamma^{0,(b)})^2
\right\}
}{
\sum_{b=1}^{B}
\left\{
(\Lambda^{0,(b)})^2
+
(\Gamma^{0,(b)})^2
\right\}
}
\right]^{1/2},\\[3pt]
\operatorname{RRMSE}_{\Gamma}
=
\left[
\frac{
\sum_{b=1}^{B}
\sum_{j=1}^p
(\hat\Gamma_j^{(b)}-\Gamma_j^{0,(b)})^2
}{
\sum_{b=1}^{B}
\sum_{j=1}^p
(\Gamma_j^{0,(b)})^2
}
\right]^{1/2},
\end{gather*}
where $b$ indexes simulation replicates. The population references $\Lambda^{0,(b)}$, $\Gamma^{0,(b)}$, and $\Gamma_1^{0,(b)},\ldots,\Gamma_p^{0,(b)}$ are described in \SM{S9}, with the covariate-specific references defined under the corresponding population MSD coupling. We also measured predictive performance using the root mean squared prediction error (RMSPE) of $\hat f_g(X_{gi})$ for $Y_{gi}$ on the samples used for MAD, pooled over the two groups. Finally, we compared the exact and approximate allocations using $\operatorname{RRMSE}_{\Gamma}$ and assessed their agreement using the Pearson correlation and the relative root mean squared difference, $\{\sum(\hat\Gamma_j^{(R)}-\hat\Gamma_j)^2/\sum\hat\Gamma_j^2\}^{1/2}$, between paired $\hat\Gamma_j^{(R)}$ and $\hat\Gamma_j$ values pooled over replicates, covariates, and covariate--response structures.\\
\indent In the coupling study, we measured three properties of each coupling $\hat\pi\in\{\hat\pi^\star,\hat\pi^{\mathrm{ind}},\hat\pi^{\varepsilon}\}$: the departure from the identity coupling by the off-diagonal mass, $1-\sum_{i=1}^n\hat\pi_{ii}$; the change in coalition values by the coalition deviation, $\{2^{-p}\sum_{H\subseteq[p]}[\hat m_{\hat\pi}(H)-\hat m_{\hat\pi}(\emptyset)]^2\}^{1/2}$, where $\hat m_{\hat\pi}(H)$ is the internal-game coalition value under $\hat\pi$, estimated from the sampled permutations for $p=20$ (\SM{S10}); and the coupling sparsity by the support size, the number of pairs $(i,i')$ with $\hat\pi_{ii'}>0$.\\
\indent In both studies, we recorded coupling-construction and allocation times.

\subsection{Simulation Results}
\label{sec:simulation-results}

\begin{figure}[p]
\centering
\includegraphics[width=\textwidth]{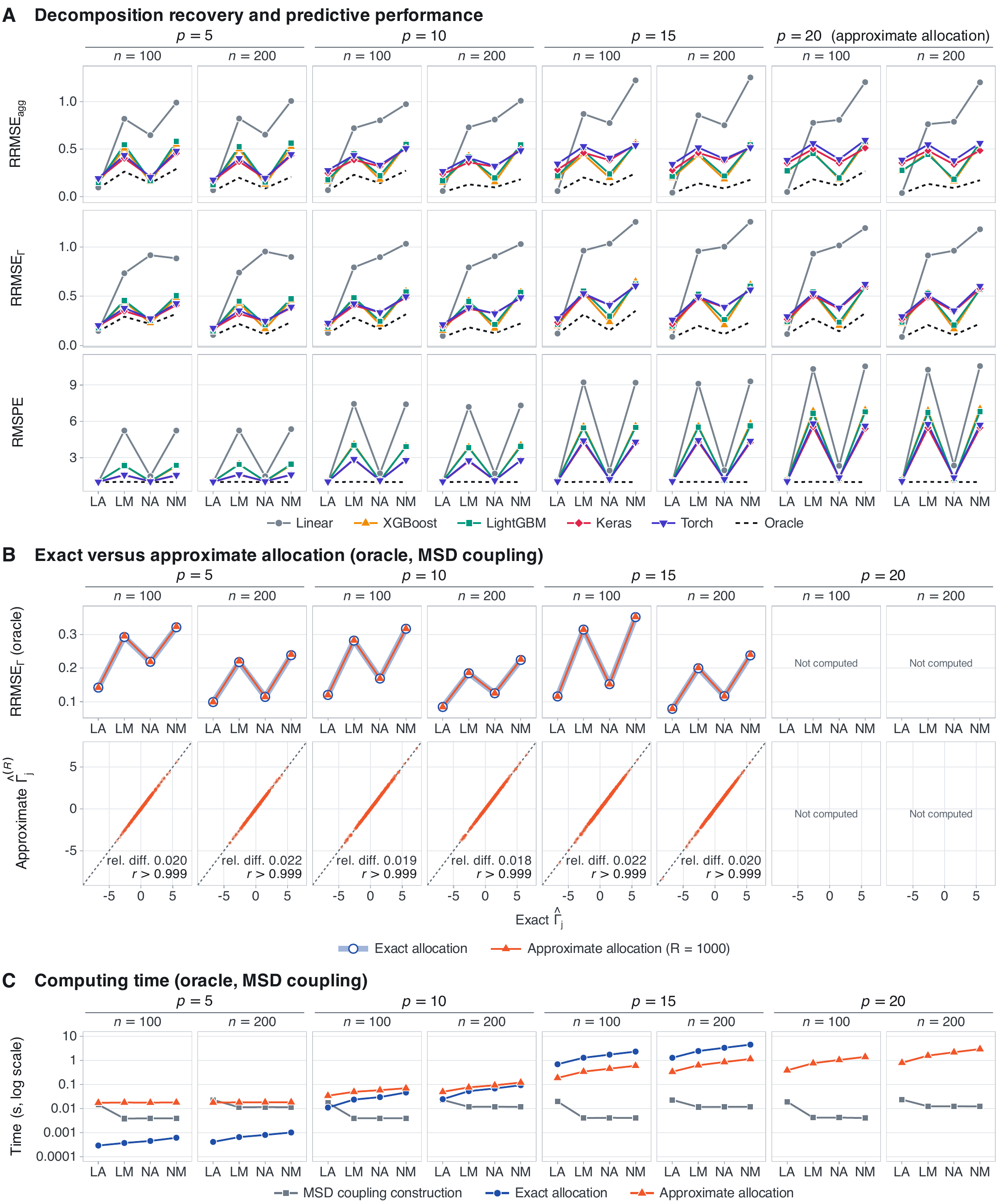}
\caption{Decomposition study. (A) $\operatorname{RRMSE}_{\mathrm{agg}}$, $\operatorname{RRMSE}_{\Gamma}$, and RMSPE of MAD with the five predictive models and the oracle benchmark under the MSD coupling, with the exact allocation for $p\in\{5,10,15\}$ and the approximate allocation for $p=20$. (B) $\operatorname{RRMSE}_{\Gamma}$ of the exact and approximate allocations (top) and approximate versus exact $\hat\Gamma_j$ pooled over the four structures, with their relative root mean squared difference (rel.\ diff.) and Pearson correlation ($r$) (bottom), under the oracle benchmark. (C) Average computing times under the oracle benchmark.}
\label{fig:sim-decomposition}
\end{figure}

\noindent \textbf{Decomposition study.}
Figure~\ref{fig:sim-decomposition}A shows the recovery of the aggregate and covariate-specific effects by MAD, measured by $\operatorname{RRMSE}_{\mathrm{agg}}$ and $\operatorname{RRMSE}_{\Gamma}$, and the predictive performance, measured by RMSPE, for the five predictive models and the oracle benchmark. As expected, the oracle benchmark, which is correctly specified in every setting and thus free of model-estimation error, attained the smallest $\operatorname{RRMSE}_{\mathrm{agg}}$, $\operatorname{RRMSE}_{\Gamma}$, and RMSPE throughout, that is, the most accurate decomposition and prediction. As $n$ increased from 100 to 200, $\operatorname{RRMSE}_{\mathrm{agg}}$ and $\operatorname{RRMSE}_{\Gamma}$ decreased consistently for the oracle benchmark, reflecting the smaller sampling error of the empirical covariate distributions.\\
\indent Under the LA structure, where linear regression is correctly specified and MAD reduces to the midpoint Blinder--Oaxaca decomposition (Proposition~\ref{prop:bo}), linear regression attained decomposition errors ($\operatorname{RRMSE}_{\mathrm{agg}}$ and $\operatorname{RRMSE}_{\Gamma}$) close to those of the oracle benchmark, whereas the machine-learning models had larger decomposition errors that tended to increase with $p$. In contrast, under the LM, NA, and NM structures, which involve nonlinear transformations or covariate interactions, linear regression had large decomposition errors, whereas the machine-learning models substantially reduced them. The prediction error (RMSPE) showed the same contrast between linear regression and the machine-learning models, indicating that decomposition accuracy depended on how well the predictive model captured the covariate--response relationship. Thus, the flexibility of MAD in the choice of predictive model translated into more accurate decompositions under nonlinear and interactive relationships.\\
\indent Figure~\ref{fig:sim-decomposition}B compares the exact and approximate allocations under the oracle benchmark for $p\in\{5,10,15\}$. Their $\operatorname{RRMSE}_{\Gamma}$ values differed by at most 0.003, and the paired $\hat\Gamma_j$ values had Pearson correlations above 0.999 and relative root mean squared differences of about 0.02. The approximate allocation, which is unbiased for the exact allocation and preserves exact adding-up (\SM{S7}), therefore closely reproduced the exact allocation.\\
\indent Figure~\ref{fig:sim-decomposition}C shows the average computing times under the oracle benchmark. Constructing the MSD coupling took less than 0.03 seconds. The exact allocation time grew rapidly with $p$, reflecting the $2^p$ coalition values (Table~\ref{tab:computational-complexity}), and reached 0.7--4.5 seconds for $p=15$. The approximate allocation time, whose leading cost scales with $Rp$ rather than $2^p$, grew more gradually, so that the approximate allocation was about four times faster than the exact allocation for $p=15$ and took at most 3 seconds for $p=20$, providing a scalable alternative for larger $p$.

\begin{figure}[p]
\centering
\includegraphics[width=\textwidth]{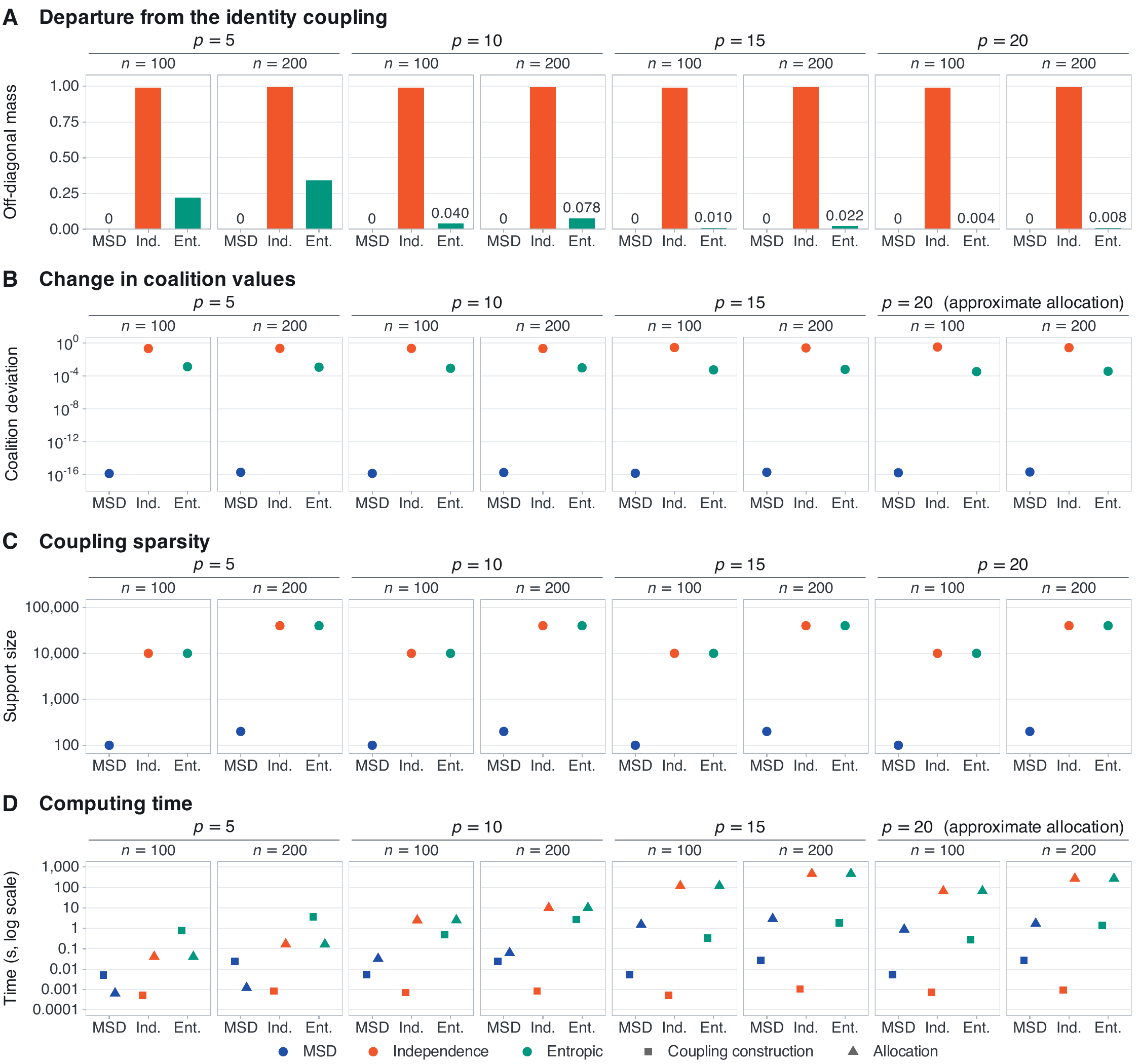}
\caption{Coupling study under the oracle benchmark with $X_A=X_B$. (A) Off-diagonal mass, (B) coalition deviation, (C) support size, and (D) average coupling-construction and allocation times of the MSD, independence (Ind.), and entropically regularized (Ent.) couplings, with the exact allocation for $p\in\{5,10,15\}$ and the approximate allocation for $p=20$. Coalition deviations and allocation times are averaged over the four covariate--response structures; for $p=20$, coalition deviations were estimated from the sampled permutations (\SM{S10}).}
\label{fig:sim-coupling}
\end{figure}

\noindent \textbf{Coupling study.}
Figure~\ref{fig:sim-coupling} summarizes the coupling study. As expected from Proposition~\ref{prop:msd-admissible}, the empirical MSD coupling recovered the identity coupling, with zero off-diagonal mass in every replicate (Figure~\ref{fig:sim-coupling}A). In contrast, the off-diagonal mass of the independence coupling was $1-1/n$, and that of the entropically regularized coupling ranged from 0.004 to 0.34, decreasing with $p$ and increasing with $n$, reflecting the violation of identity at coincidence by both alternatives (\SM{S4}). Accordingly, the coalition deviation of the MSD coupling remained at the round-off level, of the order of $10^{-16}$, whereas those of the independence and entropically regularized couplings averaged 0.21--0.32 and $3.3\times10^{-4}$--$1.3\times10^{-3}$, respectively (Figure~\ref{fig:sim-coupling}B). Hence, consistent with Proposition~\ref{prop:basic-properties}(iv), only the MSD coupling left the internal-game coalition values unchanged under partial covariate switches for every covariate--response structure when the covariate distributions coincided, whereas the two alternatives altered these values when the midpoint predictor was nonadditive (Proposition~\ref{prop:basic-properties}(ii)).\\
\indent The empirical MSD coupling had support size $n$, whereas the independence and entropically regularized couplings had full support of size $n^2$ (Figure~\ref{fig:sim-coupling}C). Constructing the MSD coupling took less than 0.03 seconds, compared with 0.3--3.6 seconds for the entropically regularized coupling, whereas the independence coupling required no optimization (Figure~\ref{fig:sim-coupling}D). Owing to its sparse support, the MSD coupling reduced the allocation time relative to the other two couplings by about two orders of magnitude, with factors of 75--78 for $n=100$ and about 160 for $n=200$ when $p\ge10$, consistent with the factor-of-$n$ reduction in Table~\ref{tab:computational-complexity}. For $p=15$ and $n=200$, for example, the allocation took 2.9 seconds under the MSD coupling and about 470 seconds under the other two couplings. Overall, the MSD coupling was the only one of the three couplings to satisfy identity at coincidence, and its sparse support also lowered allocation cost.

\section{Real Data Applications}
\label{sec:realdata}

\begin{figure}[p]
\centering
\includegraphics[width=\textwidth]{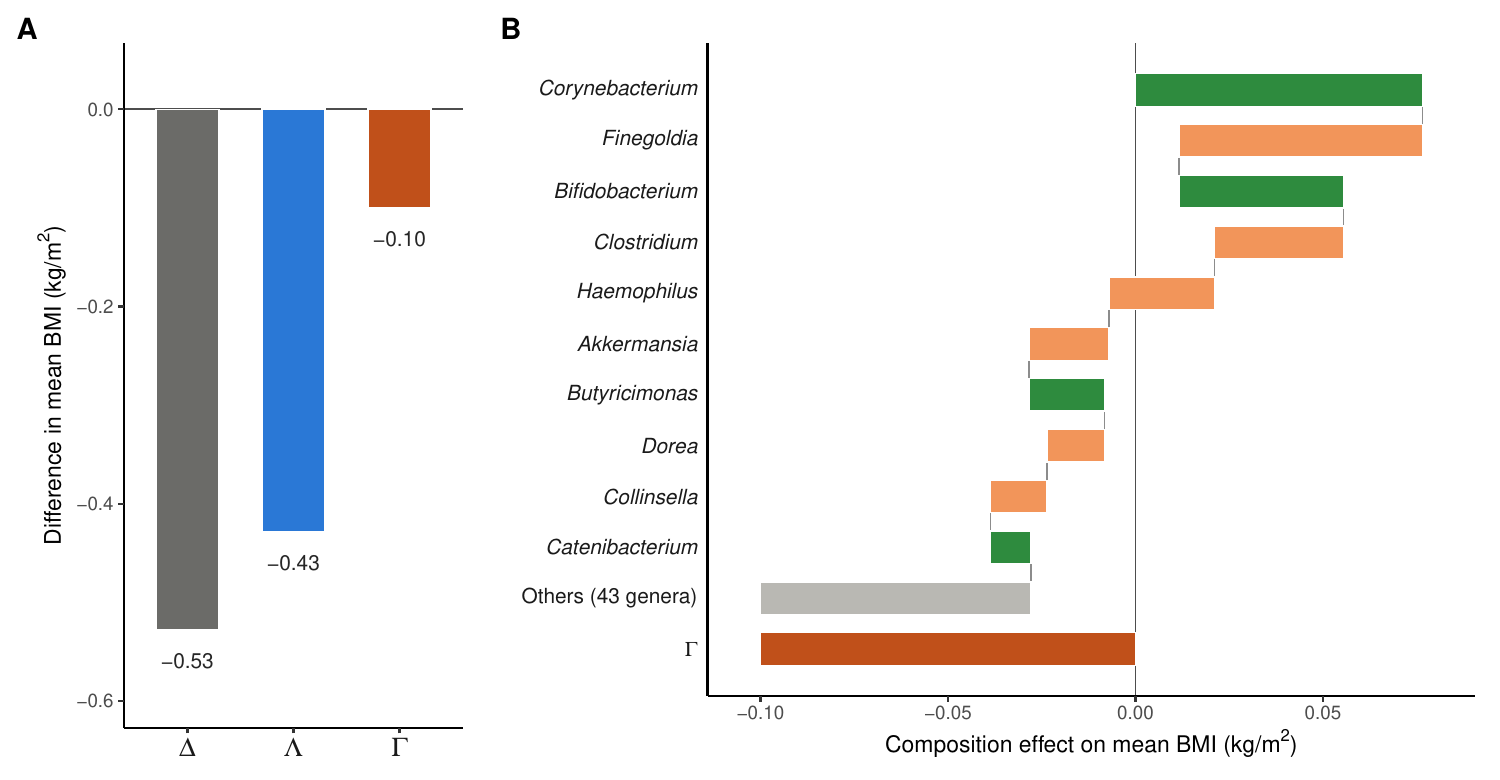}
\caption{Decomposition of the difference in mean BMI (kg/m$^2$) between UK ($n_A=730$) and US ($n_B=1{,}864$) males aged 30--70 years in the American Gut Project, based on \texttt{LightGBM} from the CLR coordinates of 53 genera, the MSD coupling, and the adaptive-budget allocation. (A) Total difference $\Delta$ (gray), aggregate structure effect $\Lambda$ (blue), and aggregate composition effect $\Gamma$ (dark orange), with $\Delta=\Lambda+\Gamma$. (B) Waterfall of the genus-specific composition effects $\Gamma_j$ of the top ten genera, followed by the sum over the remaining 43 genera (Others, gray) and $\Gamma$; positive and negative effects are shown in green and light orange, respectively. All effects are UK minus US.}
\label{fig:agp}
\end{figure}

We illustrate the proposed MAD with the MSD coupling using 16S rRNA gene amplicon data from the American Gut Project (AGP) (McDonald et al., 2018). In particular, we decompose the difference in mean body mass index (BMI, kg/m$^2$) between UK and US males aged 30--70 years, with the UK as the comparison group $A$ and the US as the baseline group $B$. We applied data quality control and transformation procedures as follows. First, we retained one fecal sample per participant with at least 10,000 reads and a BMI of at least 15 and below 60, yielding $n_A=730$ UK and $n_B=1{,}864$ US males, and the microbial genera with a mean relative abundance of at least 0.001 among all eligible participants, yielding $p=53$ genera. Second, we transformed the genus counts into centered log-ratio (CLR) coordinates (Aitchison, 1986), with zero counts replaced by 0.5, which map the compositions isometrically from the simplex onto a hyperplane of the Euclidean space (Egozcue et al., 2003). The mean BMIs of the UK and US males were 24.91 and 25.44, respectively, giving an observed difference of $-0.53$, which was significant at the 0.05 level by the Welch $t$-test ($P=0.003$; Welch, 1947).\\
\indent For each country, BMI was predicted from the CLR coordinates of the $p=53$ genera using linear regression, gradient boosting machines (\texttt{XGBoost} and \texttt{LightGBM}), and deep neural networks (\texttt{Keras} and \texttt{Torch}). \texttt{LightGBM} was selected as the predictive model because it had the smallest cross-validated root mean squared error. Because a hybrid covariate vector that combines the CLR coordinates of a UK and a US participant no longer lies on the CLR hyperplane, each predictor was evaluated at the orthogonal projection of the hybrid covariate vector onto this hyperplane, which leaves the predictions for all observed participants unchanged. We then applied the MSD coupling and the adaptive-budget allocation (Section~\ref{sec:computation}) with a tolerance of 0.0045, slightly below the target relative Monte Carlo standard error of 0.005, which required $R=646$ permutations and achieved 0.0045. Data processing, training procedures, predictive performance, and decomposition settings are detailed in \SM{S11}.\\
\indent Figure~\ref{fig:agp}A shows that, of the total difference $\Delta=-0.528$, the aggregate structure effect was $\Lambda=-0.428$ (81\%) and the aggregate composition effect was $\Gamma=-0.100$ (19\%). A substantial share of the lower mean BMI of UK males is thus attributed to differences between the two countries in how the genus composition relates to BMI, rather than to differences in the genus composition itself. Such structural differences may reflect how the host interacts with and regulates its gut microbiota, which can be governed by host factors such as genetics and physiology. Since the predictors include the genus composition alone, however, $\Lambda$ also absorbs between-country differences in other determinants of BMI that were not included as covariates, such as age, diet, living environment, and physical activity.\\
\indent Figure~\ref{fig:agp}B shows the genus-specific composition effects $\Gamma_j$ as a waterfall chart, in which the bars cumulatively add the effects of the top ten genera with the largest $|\Gamma_j|$ and then the combined effect of the remaining 43 genera to reach $\Gamma$; green and orange bars indicate composition differences associated with a higher and a lower mean BMI in the UK, respectively. The top ten genera contributed only $-0.028$ because their positive and negative effects largely offset each other, whereas the remaining 43 genera contributed $-0.072$. The two largest effects, those of \textit{Corynebacterium} ($0.076$) and \textit{Finegoldia} ($-0.065$), nearly offset each other. \textit{Bifidobacterium}, a probiotic genus familiar from yogurt and other fermented dairy products, has been inversely associated with obesity (Million et al., 2012), and \textit{Akkermansia}, a mucus-degrading genus that helps maintain the intestinal barrier, has been shown experimentally to counteract diet-induced obesity and metabolic inflammation (Everard et al., 2013). In mean relative abundance, \textit{Bifidobacterium} was slightly less abundant in the UK than in the US (1.47\% vs.\ 1.58\%), whereas \textit{Akkermansia} was more abundant in the UK than in the US (2.44\% vs.\ 2.08\%). Their composition effects had opposite signs: the positive effect of \textit{Bifidobacterium} ($\Gamma_j=0.044$) offset part of the lower mean BMI of UK males, whereas the negative effect of \textit{Akkermansia} ($\Gamma_j=-0.021$) added to it, consistent with previous findings. These genera may therefore merit investigation as targets of dietary, probiotic, or prebiotic interventions, though the substantial structure effect suggests that such interventions may need to be complemented by attention to other determinants of BMI.\\
\indent As the AGP is observational, these effects are descriptive rather than causal. On a single CPU thread, the MSD coupling took 0.5 seconds, and the adaptive-budget allocation took 18.0 minutes.
 
\section{Discussion and Conclusions}
\label{sec:discussion}

In this paper, we proposed MAD, a model-agnostic decomposition of the difference in mean response between two groups into structure and composition effects. MAD formulates a cooperative switching game with a model-switch player and individual covariate-switch players, organized into model and covariate unions. It then employs the Owen value (Owen, 1977) for a two-stage hierarchical allocation, in which the quotient game yields the aggregate structure and composition effects and the internal game allocates the aggregate composition effect across covariates. MAD thereby inherits the fair allocation properties of the Owen value, including exact two-level adding-up, and reduces to the midpoint Blinder--Oaxaca decomposition under linear predictor functions. For the hybrid covariate vectors arising in the internal game, the MSD coupling satisfies identity at coincidence, transposition equivariance, and coordinate equivariance, and has sparse support. The exact allocation is feasible for moderate $p$, and the approximate allocation based on sampled permutations provides a scalable alternative that preserves exact adding-up.\\
\indent The simulation studies supported the accuracy of MAD and the advantages of the MSD coupling. MAD with machine-learning models recovered the population structure and composition effects more accurately than MAD with linear regression under nonlinear and interactive relationships, while linear regression remained competitive under linear additive relationships, and the approximate allocation closely agreed with the exact allocation. Among the three couplings, only the MSD coupling satisfied identity at coincidence, and its sparse support substantially reduced allocation cost. The application to the American Gut Project (McDonald et al., 2018) further illustrated the practical utility of MAD: the lower mean BMI of UK males relative to US males was attributed largely to the aggregate structure effect (81\%) and partly to the aggregate composition effect (19\%), with the genus-specific effects of \textit{Bifidobacterium} and \textit{Akkermansia} consistent with previous findings. Together, these results suggest that MAD can provide accurate and interpretable decompositions with modern machine-learning models.\\
\indent As with other attribution methods based on fitted models, the effects estimated by MAD depend on how well the predictive models capture the covariate--response relationships, and, with observational data, they are descriptive rather than causal. Several extensions merit future research. First, MAD could be extended toward causal inference by combining it with appropriate identification assumptions and causal estimation methods, so that the structure and composition effects admit causal interpretations. Second, statistical inference for the estimated effects, accounting for both model estimation and sampling variability, would complement the point estimates. Third, MAD could be extended to other outcome structures, such as survival outcomes and longitudinal or time-series data, with suitable target quantities such as survival probabilities and time-indexed mean trajectories.\\
\indent Overall, MAD provides a principled and computationally practical framework for decomposing between-group differences in mean response with modern machine-learning models. Although we illustrated MAD with microbiome data, its model-agnostic formulation can be readily applied to other high-dimensional omics data, such as genomic, proteomic, and metabolomic data, and to between-group comparisons beyond biomedical research, such as in engineering and the social sciences, particularly where covariate--response relationships are complex.

\section*{Supplementary Materials}

Sections~S0--S7 of the Supplementary Materials provide proofs of Propositions~\ref{prop:owen-representation}--\ref{prop:bo}, properties of alternative couplings, and computational details of approximate allocation. Sections~S8--S10 describe the training specifications, population reference quantities, and coalition deviation estimation under approximate allocation for the simulation studies. Section~S11 details the real data application, including the genus-specific composition effects for all 53 genera.
 
\section*{Data Availability}

This study is a secondary analysis of publicly available, de-identified 16S rRNA gene amplicon data from the American Gut Project (AGP) (McDonald et al., 2018). The sequence data, sample metadata, and processed feature tables are freely available from Qiita (\url{https://qiita.ucsd.edu}; study ID 10317).

\section*{Funding}

This work was supported by the National Research Foundation of Korea (NRF) grant funded by the Korean government (MSIT) (RS-2026-25477838). 

\section*{Declaration of Competing Interest}

The author declares that he has no known competing financial interests or personal relationships that could have appeared to influence the work reported in this paper.

\section*{CRediT Authorship Contribution Statement}

\textbf{Hyunwook Koh:} Conceptualization, Methodology, Formal analysis, Investigation, Data curation, Validation, Visualization, Funding acquisition, Writing -- original draft, Writing -- review \& editing.

\newpage

\begin{table}[p]
\caption{Leading computational costs of the three empirical couplings under a unique optimal MSD assignment.}
\label{tab:computational-complexity}
\small
\renewcommand{\arraystretch}{1.15}
\begin{tabular*}{\linewidth}{@{\extracolsep{\fill}}lccc@{}}
\toprule
& MSD & Independence & Entropic \\
\midrule
Coupling construction
& $O(n^2p+n^3)$
& $O(1)$
& $O(n^2p+Tn^2)$ \\
Exact allocation
& $O(2^p n C_{\mathrm{pred}}+p2^p)$
& $O(2^p n^2 C_{\mathrm{pred}}+p2^p)$
& $O(2^p n^2 C_{\mathrm{pred}}+p2^p)$ \\
Approximate allocation
& $O(Rpn C_{\mathrm{pred}})$
& $O(Rpn^2 C_{\mathrm{pred}})$
& $O(Rpn^2 C_{\mathrm{pred}})$ \\
\bottomrule
\end{tabular*}
\vspace{3pt}
\parbox{\linewidth}{\footnotesize\textit{Note.}
$C_{\mathrm{pred}}$, cost of evaluating $\hat{\bar f}$ at one $p$-dimensional covariate vector; $T$, number of Sinkhorn iterations; $R$, number of sampled permutations.}
\end{table}

\clearpage

\begin{table}[p]
\caption{Covariate dimensions, couplings, allocation procedures, and their roles.}
\label{tab:sim-MAD-configurations}
\footnotesize
\renewcommand{\arraystretch}{1.15}
\begin{tabular*}{\linewidth}{@{\extracolsep{\fill}}lllll@{}}
\toprule
Study & Dimension & Coupling & Allocation & Role\\
\midrule
Decomposition & $p\in\{5,10,15\}$ & MSD & Exact & Primary specification\\
 & & MSD & Approximate & Allocation comparison\\
 & $p=20$ & MSD & Approximate & Primary specification\\
\midrule
Coupling & $p\in\{5,10,15\}$ & MSD, independence, entropic & Exact & Coupling comparison\\
 & $p=20$ & MSD, independence, entropic & Approximate & Coupling comparison\\
\bottomrule
\end{tabular*}
\vspace{3pt}
\parbox{\linewidth}{\footnotesize\textit{Note.}
Entropic: $\varepsilon=0.05\,\operatorname{median}_{i,i'}(C_{ii'})$ with $C_{ii'}$ in Eq.~\eqref{eq:empirical-switching-cost}; Approximate: fixed budget $R=1000$ in Eq.~\eqref{eq:composition-shapley-mc}.}
\end{table}

\clearpage

\begin{table}[p]
\caption{Four covariate--response relationship structures.}
\label{tab:sim-response-structures}
\footnotesize
\renewcommand{\arraystretch}{1.3}
\setlength{\tabcolsep}{2pt}
\resizebox{\linewidth}{!}{%
\begin{tabular}{@{}lll@{}}
\toprule
& Additive & Multiplicative\\
\midrule
Linear
&
\begin{tabular}[t]{@{}l@{}}
$\eta_A^{(\mathrm{LA})}(x)=\alpha+\sum_{j=1}^p(\beta_j+\gamma_j)x_j$\\
$\eta_B^{(\mathrm{LA})}(x)=\sum_{j=1}^p\beta_jx_j$
\end{tabular}
&
\begin{tabular}[t]{@{}l@{}}
$\eta_A^{(\mathrm{LM})}(x)=\alpha+\sum_{j=1}^p(\beta_j+\gamma_j)x_j
+\sum_{k=1}^K(\theta_k+\xi_k)\chi_{A,k}(x)$\\
$\eta_B^{(\mathrm{LM})}(x)=\sum_{j=1}^p\beta_jx_j
+\sum_{k=1}^K\theta_k\chi_{B,k}(x)$
\end{tabular}
\\
\addlinespace[6pt]
Nonlinear
&
\begin{tabular}[t]{@{}l@{}}
$\eta_A^{(\mathrm{NA})}(x)=\alpha+\sum_{j=1}^p(\beta_j+\gamma_j)\phi_{r_A(j)}(x_j)$\\
$\eta_B^{(\mathrm{NA})}(x)=\sum_{j=1}^p\beta_j\phi_{r_B(j)}(x_j)$
\end{tabular}
&
\begin{tabular}[t]{@{}l@{}}
$\eta_A^{(\mathrm{NM})}(x)=\alpha+\sum_{j=1}^p(\beta_j+\gamma_j)\phi_{r_A(j)}(x_j)
+\sum_{k=1}^K(\theta_k+\xi_k)\chi_{A,k}(x)$\\
$\eta_B^{(\mathrm{NM})}(x)=\sum_{j=1}^p\beta_j\phi_{r_B(j)}(x_j)
+\sum_{k=1}^K\theta_k\chi_{B,k}(x)$
\end{tabular}
\\
\bottomrule
\end{tabular}}
\vspace{3pt}
\parbox{\linewidth}{\footnotesize\textit{Note.}
\textit{Interaction terms.} We used $K=2p/5$ interaction terms, with two terms within each consecutive block of five covariates. For $q=1,\ldots,p/5$, group $A$ uses $\chi_{A,2q-1}(x)=x_{5q-4}x_{5q-3}$ and $\chi_{A,2q}(x)=x_{5q-2}x_{5q-1}x_{5q}$; group $B$ reverses this assignment, with $\chi_{B,2q-1}(x)=x_{5q-2}x_{5q-1}x_{5q}$ and $\chi_{B,2q}(x)=x_{5q-4}x_{5q-3}$.
\textit{Nonlinear transformations.} We used $\phi_1(t)=\sin t$, $\phi_2(t)=\cos t-e^{-1/2}$, $\phi_3(t)=\tanh t$, and $\phi_4(t)=t^2-1$, each centered to have mean zero under $N(0,1)$. Group $A$ uses the forward cycle $r_A(j)=1+\{(j-1)\bmod 4\}$; group $B$ uses the reverse cycle $r_B(j)=4-\{(j-1)\bmod 4\}$.}
\end{table}

\clearpage

\begin{table}[p]
\caption{Evaluation criteria for the two simulation studies.}
\label{tab:sim-evaluation}
\footnotesize
\renewcommand{\arraystretch}{1.15}
\begin{tabular*}{\linewidth}{@{\extracolsep{\fill}}lll@{}}
\toprule
Study & Aspect & Criterion\\
\midrule
Decomposition & Aggregate effects $(\Lambda,\Gamma)$ & $\operatorname{RRMSE}_{\mathrm{agg}}$\\
 & Covariate-specific effects $(\Gamma_1,\ldots,\Gamma_p)$ & $\operatorname{RRMSE}_{\Gamma}$\\
 & Predictive performance & RMSPE\\
 & Exact vs.\ approximate allocation & $\operatorname{RRMSE}_{\Gamma}$, Pearson correlation, relative difference\\
\midrule
Coupling & Departure from the identity coupling & Off-diagonal mass\\
 & Change in coalition values & Coalition deviation\\
 & Coupling sparsity & Support size\\
\midrule
Both & Computing time & Coupling-construction and allocation times\\
\bottomrule
\end{tabular*}
\end{table}

\newpage

\label{lastpage}
 

\begin{thebibliography}{}
 
\bibitem[Aitchison(1986)]{aitchison1986}
Aitchison, J. (1986).
\textit{The Statistical Analysis of Compositional Data}.
London: Chapman and Hall.
 
\bibitem[Audoly et al.(2025)]{audolyetal2025}
Audoly, R., McGee, R., Ocampo, S., \& Paz-Pardo, G. (2025).
A practitioner's note on the Shapley--Owen--Shorrocks decomposition.
\textit{Federal Reserve Bank of New York Staff Reports}, No. 1163.
 
\bibitem[Blinder(1973)]{blinder1973}
Blinder, A. S. (1973).
Wage discrimination: Reduced form and structural estimates.
\textit{Journal of Human Resources}, 8(4), 436--455.
 
\bibitem[Brenier(1991)]{brenier1991}
Brenier, Y. (1991).
Polar factorization and monotone rearrangement of vector-valued functions.
\textit{Communications on Pure and Applied Mathematics}, 44(4), 375--417.
 
\bibitem[Castro et al.(2009)]{castroetal2009}
Castro, J., G\'omez, D., \& Tejada, J. (2009).
Polynomial calculation of the Shapley value based on sampling.
\textit{Computers \& Operations Research}, 36(5), 1726--1730.
 
\bibitem[Chen and Guestrin(2016)]{chenguestrin2016}
Chen, T., \& Guestrin, C. (2016).
XGBoost: A scalable tree boosting system.
In \textit{Proceedings of the 22nd ACM SIGKDD International Conference on
Knowledge Discovery and Data Mining} (pp. 785--794).
 
\bibitem[Chollet et al.(2015)]{cholletetal2015}
Chollet, F., et al. (2015).
Keras.
\url{https://keras.io}.
 
\bibitem[Cominetti and San Mart\'in(1994)]{cominettisanmartin1994}
Cominetti, R., \& San Mart\'in, J. (1994).
Asymptotic analysis of the exponential penalty trajectory in linear programming.
\textit{Mathematical Programming}, 67(1--3), 169--187.
 
\bibitem[Cotton(1988)]{cotton1988}
Cotton, J. (1988).
On the decomposition of wage differentials.
\textit{The Review of Economics and Statistics}, 70(2), 236--243.
 
\bibitem[Cuturi(2013)]{cuturi2013}
Cuturi, M. (2013).
Sinkhorn distances: Lightspeed computation of optimal transport.
In \textit{Advances in Neural Information Processing Systems} (Vol. 26, pp. 2292--2300).
 
\bibitem[Egozcue et al.(2003)]{egozcueetal2003}
Egozcue, J. J., Pawlowsky-Glahn, V., Mateu-Figueras, G., \& Barcel\'o-Vidal, C. (2003).
Isometric logratio transformations for compositional data analysis.
\textit{Mathematical Geology}, 35(3), 279--300.
 
\bibitem[Everard et al.(2013)]{everardetal2013}
Everard, A., Belzer, C., Geurts, L., Ouwerkerk, J. P., Druart, C., et al. (2013).
Cross-talk between \textit{Akkermansia muciniphila} and intestinal epithelium controls diet-induced obesity.
\textit{Proceedings of the National Academy of Sciences of the United States of America}, 110(22), 9066--9071.

\bibitem[Friedman(2001)]{friedman2001}
Friedman, J. H. (2001).
Greedy function approximation: A gradient boosting machine.
\textit{The Annals of Statistics}, 29(5), 1189--1232.
 
\bibitem[Ke et al.(2017)]{keetal2017}
Ke, G., Meng, Q., Finley, T., Wang, T., Chen, W., Ma, W., Ye, Q., \& Liu, T.-Y. (2017).
LightGBM: A highly efficient gradient boosting decision tree.
In \textit{Advances in Neural Information Processing Systems} (Vol. 30, pp. 3146--3154).
 
\bibitem[Kuhn(1955)]{kuhn1955}
Kuhn, H. W. (1955).
The Hungarian method for the assignment problem.
\textit{Naval Research Logistics Quarterly}, 2(1--2), 83--97.
 
\bibitem[LeCun et al.(2015)]{lecunetal2015}
LeCun, Y., Bengio, Y., \& Hinton, G. (2015).
Deep learning.
\textit{Nature}, 521, 436--444.
 
\bibitem[Mao et al.(2025)]{maoetal2025}
Mao, M., Raiola, A., \& Yang, D. (2025).
Double machine learning for Oaxaca--Blinder decomposition.
\textit{Economics Letters}, 255, 112525.
 
\bibitem[McCann(1995)]{mccann1995}
McCann, R. J. (1995).
Existence and uniqueness of monotone measure-preserving maps.
\textit{Duke Mathematical Journal}, 80(2), 309--323.
 
\bibitem[McDonald et al.(2018)]{mcdonaldetal2018}
McDonald, D., Hyde, E., Debelius, J. W., Morton, J. T., Gonzalez, A., et al. (2018).
American Gut: An open platform for citizen science microbiome research.
\textit{mSystems}, 3(3), e00031-18.
 
\bibitem[Million et al.(2012)]{millionetal2012}
Million, M., Maraninchi, M., Henry, M., Armougom, F., Richet, H., et al. (2012).
Obesity-associated gut microbiota is enriched in \textit{Lactobacillus reuteri} and depleted in \textit{Bifidobacterium animalis} and \textit{Methanobrevibacter smithii}.
\textit{International Journal of Obesity}, 36(6), 817--825.
 
\bibitem[Mitchell et al.(2022)]{mitchelletal2022}
Mitchell, R., Cooper, J., Frank, E., \& Holmes, G. (2022).
Sampling permutations for Shapley value estimation.
\textit{Journal of Machine Learning Research}, 23(43), 1--46.
 
\bibitem[Neumark(1988)]{neumark1988}
Neumark, D. (1988).
Employers' discriminatory behavior and the estimation of wage discrimination.
\textit{Journal of Human Resources}, 23(3), 279--295.
 
\bibitem[Oaxaca(1973)]{oaxaca1973}
Oaxaca, R. L. (1973).
Male-female wage differentials in urban labor markets.
\textit{International Economic Review}, 14(3), 693--709.
 
\bibitem[Owen(1977)]{owen1977}
Owen, G. (1977).
Values of games with a priori unions.
In R. Henn \& O. Moeschlin (Eds.),
\textit{Mathematical Economics and Game Theory: Essays in Honor of Oskar Morgenstern}
(Lecture Notes in Economics and Mathematical Systems, Vol. 141, pp. 76--88).
Berlin: Springer.
 
\bibitem[Paszke et al.(2019)]{paszkeetal2019}
Paszke, A., Gross, S., Massa, F., Lerer, A., Bradbury, J., Chanan, G., Killeen, T., Lin, Z., Gimelshein, N., Antiga, L., Desmaison, A., K\"opf, A., Yang, E., DeVito, Z., Raison, M., Tejani, A., Chilamkurthy, S., Steiner, B., Fang, L., Bai, J., \& Chintala, S. (2019).
PyTorch: An imperative style, high-performance deep learning library.
In \textit{Advances in Neural Information Processing Systems} (Vol. 32, pp. 8024--8035).
 
\bibitem[Peyr\'e and Cuturi(2019)]{peyrecuturi2019}
Peyr\'e, G., \& Cuturi, M. (2019).
Computational optimal transport: With applications to data science.
\textit{Foundations and Trends in Machine Learning}, 11(5--6), 355--607.
 
\bibitem[Reimers(1983)]{reimers1983}
Reimers, C. W. (1983).
Labor market discrimination against Hispanic and Black men.
\textit{The Review of Economics and Statistics}, 65(4), 570--579.
 
\bibitem[Rothe(2015)]{rothe2015}
Rothe, C. (2015).
Decomposing the composition effect: The role of covariates in determining between-group differences in economic outcomes.
\textit{Journal of Business \& Economic Statistics}, 33(3), 323--337.
 
\bibitem[Shapley(1953)]{shapley1953}
Shapley, L. S. (1953).
A value for $n$-person games.
In H. W. Kuhn \& A. W. Tucker (Eds.),
\textit{Contributions to the Theory of Games II} (pp. 307--317).
Princeton, NJ: Princeton University Press.
 
\bibitem[Shorrocks(2013)]{shorrocks2013}
Shorrocks, A. F. (2013).
Decomposition procedures for distributional analysis: A unified framework based on the Shapley value.
\textit{The Journal of Economic Inequality}, 11(1), 99--126.
 
\bibitem[Singh et al.(2024)]{singhetal2024}
Singh, H., Xia, F., Subbaswamy, A., Gossmann, A., \& Feng, J. (2024).
A hierarchical decomposition for explaining ML performance discrepancies.
In \textit{Advances in Neural Information Processing Systems} (Vol. 37, pp. 128516--128555).
 
\bibitem[Sinning et al.(2008)]{sinningetal2008}
Sinning, M., Hahn, M., \& Bauer, T. K. (2008).
The Blinder--Oaxaca decomposition for nonlinear regression models.
\textit{The Stata Journal}, 8(4), 480--492.
 
\bibitem[Villani(2009)]{villani2009}
Villani, C. (2009).
\textit{Optimal Transport: Old and New}.
Berlin: Springer.
 
\bibitem[Welch(1947)]{welch1947}
Welch, B. L. (1947).
The generalization of `Student's' problem when several different population variances are involved.
\textit{Biometrika}, 34(1--2), 28--35.
 
\bibitem[Yun(2004)]{yun2004}
Yun, M.-S. (2004).
Decomposing differences in the first moment.
\textit{Economics Letters}, 82(2), 275--280.
 
\bibitem[Zhang et al.(2023)]{zhangetal2023}
Zhang, H., Singh, H., Ghassemi, M., \& Joshi, S. (2023).
``Why did the model fail?'': Attributing model performance changes to distribution shifts.
In \textit{Proceedings of the 40th International Conference on Machine Learning},
\textit{Proceedings of Machine Learning Research}, 202, 41550--41578.
 
\end{thebibliography}
\end{document}


\maketitle

\section*{Overview}

\indent This supplement provides proofs and technical details for MAD. Sections~S0--S7 accompany the Methods section, covering the notation, the Owen representation, the comparison with a flat Shapley allocation, the MSD coupling and alternative couplings, further properties, the Blinder--Oaxaca reduction, and approximate allocation and computational complexity. Sections~S8--S10 give the simulation training specifications, population reference quantities, and coalition-deviation calculation, and Section~S11 gives the details of the real data application.\\

\section*{S0. Notation}

\indent We follow the notation of the main text and collect here the quantities used in the proofs. We let $[p]:=\{1,\dots,p\}$ and, for $H\subseteq[p]$, $H^c:=[p]\setminus H$. For group $g\in\{A,B\}$, we let $P_g$ denote the distribution of $X_g\in\mathbb{R}^p$, with mean vector $\mu_{X_g}$ and covariance matrix $\Sigma_{X_g}$, and $P_{g,H}$ the distribution of the subvector $X_{g,H}$, with $P_{g,j}:=P_{g,\{j\}}$. We let $f_g:\mathbb{R}^p\to\mathbb{R}$ denote the group-specific predictor function, $f_g(x)=\mathbb{E}(Y_g\mid X_g=x)$, so that $\mu_{Y_g}=\mathbb{E}\{f_g(X_g)\}$ and $\Delta=\mu_{Y_A}-\mu_{Y_B}$. All expectations used below are assumed finite.\\
\indent For $H\subseteq[p]$, the hybrid covariate vector is
\begin{equation*}
\bigl(X_{\Omega(H)}\bigr)_j
:=
\begin{cases}
X_{Aj}, & j\in H,\\
X_{Bj}, & j\in H^c,
\end{cases}
\qquad j\in[p],
\end{equation*}
so that $X_{\Omega(\emptyset)}=X_B$ and $X_{\Omega([p])}=X_A$. A coupling $\pi$ of $P_A$ and $P_B$ is a joint distribution of $(X_A,X_B)$ on $\mathbb{R}^p\times\mathbb{R}^p$ with marginals $P_A$ and $P_B$, respectively, and $Q_{\Omega(H)}^\pi:=\mathcal{L}_\pi(X_{\Omega(H)})$ denotes the distribution of the hybrid covariate vector under $\pi$. The marginal constraints of the coupling give
\begin{equation*}
\mathcal{L}_{\pi}\bigl((X_{\Omega(H)})_H\bigr)=P_{A,H},
\qquad
\mathcal{L}_{\pi}\bigl((X_{\Omega(H)})_{H^c}\bigr)=P_{B,H^c},
\end{equation*}
and, at the two endpoints,
\begin{equation}\label{eq:s-endpoints}
Q_{\Omega(\emptyset)}^\pi=P_B,
\qquad
Q_{\Omega([p])}^\pi=P_A
\end{equation}
for any coupling $\pi$.\\
\indent The cooperative switching game has player set and \textit{a priori} union structure
\begin{equation*}
N=\{M,C_1,\dots,C_p\},
\qquad
U_M:=\{M\},
\qquad
U_C:=\{C_1,\dots,C_p\},
\qquad
\mathcal{P}:=\{U_M,U_C\}.
\end{equation*}
For $S\subseteq N$, we let $H(S):=\{j\in[p]:C_j\in S\}$, and $g(S):=A$ if $M\in S$ and $g(S):=B$ otherwise. Given $\pi$, the characteristic function is
\begin{equation*}
v^\pi(S)
:=
\mathbb{E}_{\pi}\bigl\{f_{g(S)}(X_{\Omega(H(S))})\bigr\}
-\mu_{Y_B},
\qquad S\subseteq N.
\end{equation*}
By \eqref{eq:s-endpoints},
\begin{equation}\label{eq:s-game-endpoints}
v^\pi(\emptyset)=0,
\quad
v^\pi(U_M)=\mathbb{E}\{f_A(X_B)\}-\mu_{Y_B},
\quad
v^\pi(U_C)=\mathbb{E}\{f_B(X_A)\}-\mu_{Y_B},
\quad
v^\pi(N)=\Delta
\end{equation}
for any coupling $\pi$. With
\begin{equation*}
d(x):=f_A(x)-f_B(x),
\qquad
\bar f(x):=\tfrac12\{f_A(x)+f_B(x)\},
\end{equation*}
the aggregate structure and composition effects, the payoffs of $U_M$ and $U_C$ in the quotient game, are
\begin{equation}\label{eq:s-aggregate}
\Lambda
=
\tfrac12\mathbb{E}_{P_A}\{d(X)\}
+
\tfrac12\mathbb{E}_{P_B}\{d(X)\},
\qquad
\Gamma
=
\mathbb{E}_{P_A}\{\bar f(X)\}
-
\mathbb{E}_{P_B}\{\bar f(X)\},
\end{equation}
with $\Lambda+\Gamma=\Delta$. Neither quantity depends on $\pi$.\\
\indent For $H\subseteq[p]$, we let $\mathcal{C}_H:=\{C_j:j\in H\}$, and we abbreviate the Shapley weights as
\begin{equation*}
\kappa_k:=\frac{k!\,(p-k-1)!}{p!},
\qquad k=0,\dots,p-1.
\end{equation*}
Given $\pi$, the covariate-specific composition effect of covariate $j$ is
\begin{equation}\label{eq:s-gamma-j}
\begin{aligned}
\Gamma_j^\pi
&=
\sum_{H\subseteq[p]\setminus\{j\}}
\kappa_{|H|}\,\frac12
\Bigl[
v^\pi(\mathcal{C}_{H\cup\{j\}})-v^\pi(\mathcal{C}_H)\\
&\hspace{39mm}
+
v^\pi(U_M\cup\mathcal{C}_{H\cup\{j\}})
-v^\pi(U_M\cup\mathcal{C}_H)
\Bigr]\\
&=
\sum_{H\subseteq[p]\setminus\{j\}}
\kappa_{|H|}
\left[
\mathbb{E}_{\pi}\{\bar f(X_{\Omega(H\cup\{j\})})\}
-
\mathbb{E}_{\pi}\{\bar f(X_{\Omega(H)})\}
\right].
\end{aligned}
\end{equation}
MAD uses the MSD coupling $\pi^\star$ of Section~S3, with $\Gamma_j:=\Gamma_j^{\pi^\star}$.\\

\section*{S1. Proof of the Owen Representation and Its Consequences}

\indent For a game $(N,v)$ with \textit{a priori} unions $\mathcal{P}=\{U_1,\dots,U_m\}$ and $i\in U_k$, we let $n_k:=|U_k|$ and $U_R:=\bigcup_{\ell\in R}U_\ell$ for $R\subseteq\{1,\dots,m\}$. The Owen value \citep{owen1977} is
\begin{equation}\label{eq:s-owen}
\operatorname{Ow}_i(N,v,\mathcal{P})
=
\sum_{R\subseteq\{1,\dots,m\}\setminus\{k\}}
\sum_{T\subseteq U_k\setminus\{i\}}
\frac{|R|!\,(m-|R|-1)!}{m!}
\frac{|T|!\,(n_k-|T|-1)!}{n_k!}
\{v(U_R\cup T\cup\{i\})-v(U_R\cup T)\}.
\end{equation}

\mtresult{Proposition 1 (Owen representation and its consequences).}
\textit{For any coupling $\pi$, the Owen allocation of $(N,v^\pi)$ under $\mathcal{P}$ assigns $\Lambda$ to $M$ and $\Gamma_j^\pi$ to $C_j$, for $j\in[p]$, and satisfies: (i) two-level coherence, $\sum_{j=1}^p\Gamma_j^\pi=\Gamma$; (ii) efficiency, $\Delta=\Lambda+\sum_{j=1}^p\Gamma_j^\pi$; (iii) the null-player property: if $v^\pi(S\cup\{C_j\})=v^\pi(S)$ for every $S\subseteq N\setminus\{C_j\}$, then $\Gamma_j^\pi=0$; (iv) symmetry within the covariate union: if $v^\pi(S\cup\{C_j\})=v^\pi(S\cup\{C_k\})$ for every $S\subseteq N\setminus\{C_j,C_k\}$, then $\Gamma_j^\pi=\Gamma_k^\pi$; and (v) linearity: if $v^\pi=a_1v'+a_2v''$, then each payoff is the corresponding linear combination of the Owen payoffs under $v'$ and $v''$ with the same union structure $\mathcal{P}$.}

\begin{proof}
\indent Because $\mathcal{P}$ contains two unions and $U_M$ is a singleton,
\begin{equation*}
\operatorname{Ow}_M
=
\tfrac12\{v^\pi(U_M)-v^\pi(\emptyset)\}
+
\tfrac12\{v^\pi(N)-v^\pi(U_C)\}
=
\Lambda
\end{equation*}
by \eqref{eq:s-game-endpoints}--\eqref{eq:s-aggregate}. The quotient-game payoff of $U_C$ is similarly $\Gamma$, and Shapley efficiency gives $\Lambda+\Gamma=\Delta$.\\
\indent For $C_j$, the two possible orders of the unions each have weight $1/2$, while each coalition $\mathcal{C}_H$ of covariate-switch players preceding $C_j$ within $U_C$ has weight $\kappa_{|H|}$. Substituting these weights into \eqref{eq:s-owen} gives the first equality in \eqref{eq:s-gamma-j}, so $\operatorname{Ow}_{C_j}=\Gamma_j^\pi$. Because $v^\pi(\mathcal{C}_H)$ and $v^\pi(U_M\cup\mathcal{C}_H)$ evaluate $f_B$ and $f_A$, respectively, at $X_{\Omega(H)}$, averaging the two marginal contributions for each $H$ gives the second equality in \eqref{eq:s-gamma-j}. This is the ordinary Shapley formula \citep{shapley1953} for the set function
\[
H\longmapsto \mathbb{E}_{\pi}\{\bar f(X_{\Omega(H)})\}.
\]
Therefore, by Shapley efficiency and \eqref{eq:s-endpoints},
\begin{equation}\label{eq:s-internal-efficiency}
\sum_{j=1}^p\Gamma_j^\pi
=
\mathbb{E}_{P_A}\{\bar f(X)\}
-
\mathbb{E}_{P_B}\{\bar f(X)\}
=
\Gamma.
\end{equation}
Equation~\eqref{eq:s-internal-efficiency} proves two-level coherence, and combining it with $\Lambda+\Gamma=\Delta$ proves efficiency.\\
\indent The null-player, within-union symmetry, and linearity properties follow directly from the corresponding properties of the Owen value, or equivalently from \eqref{eq:s-owen}.
\end{proof}

\section*{S2. Formal Comparison with a Flat Shapley Allocation}

\indent Given $\pi$, we define
\[
\Lambda^{\mathrm{flat}}(\pi)
:=
\operatorname{Sh}_M(N,v^\pi),
\qquad
\Gamma_j^{\mathrm{flat}}(\pi)
:=
\operatorname{Sh}_{C_j}(N,v^\pi)
\]
as the Shapley payoffs \citep{shapley1953} under a flat allocation on the full player set $N$ that ignores the \textit{a priori} union structure. The model-switch payoff is
\begin{equation*}
\Lambda^{\mathrm{flat}}(\pi)
=
\sum_{H\subseteq[p]}
\omega_{|H|}
\mathbb{E}_{\pi}\{d(X_{\Omega(H)})\},
\qquad
\omega_k:=\frac{k!\,(p-k)!}{(p+1)!}.
\end{equation*}
Thus, the flat payoff averages model-switch contributions over both endpoint and intermediate hybrid covariate vectors, whereas $\Lambda$ in \eqref{eq:s-aggregate} uses only the two endpoints. Consequently, $\Lambda^{\mathrm{flat}}(\pi)$ and $\Lambda$ need not coincide. Ordinary Shapley efficiency further gives
\begin{equation}\label{eq:s-flat-efficiency}
\sum_{j=1}^p\Gamma_j^{\mathrm{flat}}(\pi)
=
\Delta-\Lambda^{\mathrm{flat}}(\pi),
\end{equation}
so, whenever $\Lambda^{\mathrm{flat}}(\pi)\neq\Lambda$, the flat covariate-switch payoffs do not sum to $\Gamma=\Delta-\Lambda$.\\
\indent A simple example shows that this discrepancy can occur. We let $p=2$, take $P_B$ and $P_A$ to be point masses at $(0,0)$ and $(1,1)$, respectively, and let $f_B\equiv0$ and $f_A(x)=x_1x_2$, for which the coupling is unique. Then $\Delta=1$ and
\[
\Lambda=\Gamma=\tfrac12.
\]
In the flat three-player game, the model-switch marginal contribution is zero at the hybrid covariate vectors $(0,0)$, $(1,0)$, and $(0,1)$ and equals one at $(1,1)$. Since the Shapley weight of the predecessor coalition $\{C_1,C_2\}$ is $1/3$,
\[
\Lambda^{\mathrm{flat}}=\tfrac13\neq\tfrac12=\Lambda,
\]
and \eqref{eq:s-flat-efficiency} gives
\[
\Gamma_1^{\mathrm{flat}}+\Gamma_2^{\mathrm{flat}}
=
\tfrac23
\neq
\tfrac12
=
\Gamma.
\]
Thus, a flat Shapley allocation need not preserve the aggregate structure--composition split.\\

\section*{S3. Existence, Uniqueness, and Admissibility of the MSD Coupling}

\indent Because $P_A$ and $P_B$ have finite second moments, we define
\begin{equation}\label{eq:s-scale-cost}
s_j^2
:=
\tfrac12\{(\Sigma_{X_A})_{jj}+(\Sigma_{X_B})_{jj}\},
\qquad
J:=\{j\in[p]:s_j>0\},
\end{equation}
and
\begin{equation}\label{eq:s-cost}
c(x_A,x_B)
:=
\sum_{j\in J}
\left(\frac{x_{Aj}-x_{Bj}}{s_j}\right)^2.
\end{equation}
The MSD coupling is a minimizer of the expected switching cost,
\begin{equation}\label{eq:s-msd}
\pi^\star
\in
\operatorname*{arg\,min}_{\pi}
\mathbb{E}_{\pi}\{c(X_A,X_B)\},
\end{equation}
where the minimum is taken over all couplings $\pi$ of $P_A$ and $P_B$, and multiple minimizers, if any, are resolved by a prespecified deterministic tie-breaking rule. For $j\notin J$, $X_{Aj}$ and $X_{Bj}$ are almost surely constant, so a coupling of $P_A$ and $P_B$ is determined by its restriction to the coordinates in $J$. After standardizing the coordinates in $J$ by $x_j\mapsto x_j/s_j$, \eqref{eq:s-msd} is equivalent to the Kantorovich optimal-transport problem on $\mathbb{R}^{|J|}$ with squared Euclidean ground cost, and finite second moments ensure the existence of a minimizer \citep{villani2009,santambrogio2015,peyrecuturi2019}. A standard sufficient condition for uniqueness is absolute continuity, with respect to Lebesgue measure on $\mathbb{R}^{|J|}$, of either standardized endpoint distribution restricted to the coordinates in $J$ \citep{brenier1991,mccann1995}.\\

\mtresult{Proposition 2 (Admissibility).}
\textit{On any domain where the MSD minimizer in \eqref{eq:s-msd} is unique, the coupling rule $(P_A,P_B)\mapsto\pi^\star$ is admissible. Under nonuniqueness, admissibility holds if the tie-breaking rule is equivariant to group exchange and common covariate relabeling.}

\begin{proof}
\indent Suppose first that $P_A=P_B=P$. The identity coupling $\mathcal{L}(X,X)$ has zero cost, so any minimizer has zero cost, which implies $X_{Aj}=X_{Bj}$ almost surely for every $j\in J$. For $j\notin J$, the common marginal coordinate is constant. Hence $X_A=X_B$ almost surely, and the identity coupling is the unique minimizer, so identity at coincidence holds without tie-breaking.\\
\indent We let $\tau(x_A,x_B):=(x_B,x_A)$. Because \eqref{eq:s-scale-cost}--\eqref{eq:s-cost} are invariant to group exchange, $\pi\mapsto\tau_{\#}\pi$ preserves the objective and maps the coupling set for $(P_A,P_B)$ bijectively onto that for $(P_B,P_A)$. Thus, unique minimizers are transposed under group exchange, and the same holds under nonuniqueness when the tie-breaking rule is transposition equivariant. Likewise, a common permutation of the coordinates maps feasible couplings bijectively to the relabeled problem and leaves the standardized cost unchanged, so unique minimizers, or minimizers chosen by a coordinate-equivariant tie-breaking rule, are relabeled in the same way. These are the three admissibility requirements.
\end{proof}

\indent The maximum-entropy tie-breaking rule of the empirical implementation, given in the Computational Procedures of the main text, satisfies this equivariance requirement. Exchanging the groups transposes the cost matrix $(C_{ii'})$ and hence the set $\widehat{\Pi}_{\mathrm{MSD}}$ of empirical minimizers, and the entropy $\mathcal{H}$ is invariant under transposition, so the selected minimizer $\hat\pi^\star$ is transposed. A common relabeling of the covariates leaves the cost matrix, and hence $\hat\pi^\star$, unchanged.\\

\section*{S4. Independence and Entropically Regularized Couplings}

\indent Under the independence coupling $\pi^{\mathrm{ind}}:=P_A\otimes P_B$,
\begin{equation*}
Q_{\Omega(H)}^{\pi^{\mathrm{ind}}}
=
P_{A,H}\otimes P_{B,H^c},
\qquad H\subseteq[p],
\end{equation*}
so the switched and retained blocks are independent, whereas the dependence within each block is preserved. For $\varepsilon>0$, the entropically regularized coupling is
\begin{equation*}
\pi^\varepsilon
:=
\operatorname*{arg\,min}_{\pi}
\bigl[
\mathbb{E}_\pi\{c(X_A,X_B)\}
+
\varepsilon\,\mathrm{KL}(\pi\,\|\,P_A\otimes P_B)
\bigr],
\end{equation*}
where $\mathrm{KL}$ denotes the Kullback--Leibler divergence and the minimum is taken over all couplings of $P_A$ and $P_B$. Under the stated finite-second-moment conditions, the entropically regularized problem has a unique minimizer \citep{csiszar1975,peyrecuturi2019}.\\

\mtresult{Lemma S1.}
\textit{The independence and entropically regularized couplings satisfy conditions (ii) and (iii) of Definition~1 of the main text but violate condition (i): if $P_A=P_B=P$ and $P$ is not a point mass, neither equals the identity coupling $\pi^{\mathrm{id}}:=\mathcal{L}(X,X)$ with $X\sim P$. Hence, neither coupling rule is admissible on a domain containing such a pair $(P,P)$.}

\begin{proof}
\indent Exchanging $P_A$ and $P_B$ transposes $P_A\otimes P_B$, and a common relabeling of the coordinates relabels it in the same way. Both the switching cost $c$ and $\mathrm{KL}(\cdot\,\|\,P_A\otimes P_B)$ are invariant under these operations, and $\pi^\varepsilon$ is unique, so $\pi^\varepsilon$ is transposed and relabeled in the same way.\\
\indent We now let $P_A=P_B=P$ and $\mathcal{I}:=\{(x,x):x\in\mathbb{R}^p\}$, so that $\pi^{\mathrm{id}}(\mathcal{I})=1$. If $P$ is not a point mass, then $q:=(P\otimes P)(\mathcal{I}^c)>0$, so $\pi^{\mathrm{ind}}=P\otimes P\neq\pi^{\mathrm{id}}$.\\
\indent Suppose that $\pi^\varepsilon=\pi^{\mathrm{id}}$. If $\mathrm{KL}(\pi^{\mathrm{id}}\,\|\,P\otimes P)=\infty$, the objective is infinite at $\pi^{\mathrm{id}}$ but finite at $P\otimes P$, a contradiction. Otherwise, $\pi^{\mathrm{id}}\ll P\otimes P$, so $P=\sum_x p_x\delta_x$ is purely atomic and $\mathrm{KL}(\pi^{\mathrm{id}}\,\|\,P\otimes P)=-\sum_x p_x\log p_x<\infty$. For $t\in(0,1)$, we let $\pi_t:=(1-t)\pi^{\mathrm{id}}+t\,P\otimes P$. Because $c=0$ on $\mathcal{I}$, $\mathbb{E}_{\pi_t}\{c(X_A,X_B)\}=t\,\mathbb{E}_{P\otimes P}\{c(X_A,X_B)\}$. On $\mathcal{I}^c$, $\pi_t=t\,P\otimes P$, which contributes $qt\log t$ to $\mathrm{KL}(\pi_t\,\|\,P\otimes P)$, whereas the contribution of $\mathcal{I}$ differs from its value at $t=0$ by $O(t)$ because $\mathrm{KL}(\pi^{\mathrm{id}}\,\|\,P\otimes P)<\infty$. Hence, the objective at $\pi_t$ equals its value at $\pi^{\mathrm{id}}$ plus an increment $\varepsilon qt\log t+O(t)$, which is negative for sufficiently small $t$, contradicting the optimality of $\pi^{\mathrm{id}}$.
\end{proof}

\indent The consequences of this violation at coincident covariate distributions are stated in Proposition~3(iv) and proved in Section~S5.\\

\section*{S5. Proof of the Further Properties of MAD}

\mtresult{Proposition 3 (Further properties of MAD).}
\textit{MAD satisfies the following properties: (i) coupling invariance of the aggregate effects: $\Lambda$ and $\Gamma$ do not depend on the coupling; (ii) coupling invariance under an additive midpoint predictor: if $\bar f(x)=c_0+\sum_{j=1}^p\psi_j(x_j)$, where $c_0$ is constant and the univariate functions $\psi_1,\dots,\psi_p$ have finite expectations under $P_A$ and $P_B$, then $\Gamma_j^\pi=\mathbb{E}_{P_A}\{\psi_j(X_j)\}-\mathbb{E}_{P_B}\{\psi_j(X_j)\}$ for any coupling $\pi$ and $j\in[p]$; (iii) null structure effect: if $f_A(x)=f_B(x)$ for $P_A$- and $P_B$-almost every $x$, then $\Lambda=0$; (iv) null composition effects at coincident distributions: if $P_A=P_B=P$, then $\Gamma=0$ for any coupling $\pi$; if the coupling is selected by a rule satisfying condition (ii) of Definition~1 of the main text, then $\Gamma_j^\pi=0$ for every $j\in[p]$; if the rule satisfies condition (i), then $\mathbb{E}_\pi\{\bar f(X_{\Omega(H)})\}=\mathbb{E}_P\{\bar f(X)\}$ for every $H\subseteq[p]$; and (v) group-exchange equivariance: if the coupling is selected by a rule satisfying condition (ii), exchanging groups $A$ and $B$ reverses the signs of $\Lambda$, $\Gamma$, and each covariate-specific composition effect.}

\begin{proof}
\indent Part (i) follows immediately from \eqref{eq:s-aggregate}, which depends only on the endpoint predictor functions and endpoint covariate distributions.\\
\indent For part (ii), additivity gives, for $H\subseteq[p]\setminus\{j\}$,
\[
\mathbb{E}_{\pi}\{\bar f(X_{\Omega(H\cup\{j\})})\}
-
\mathbb{E}_{\pi}\{\bar f(X_{\Omega(H)})\}
=
\mathbb{E}_{P_A}\{\psi_j(X_j)\}
-
\mathbb{E}_{P_B}\{\psi_j(X_j)\},
\]
because the two hybrid vectors differ only in coordinate $j$, whose distribution is $P_{A,j}$ under the first and $P_{B,j}$ under the second. The right-hand side does not depend on $H$ or $\pi$, and substitution into \eqref{eq:s-gamma-j}, together with $\sum_{H\subseteq[p]\setminus\{j\}}\kappa_{|H|}=1$, proves part (ii).\\
\indent For part (iii), if $f_A(x)=f_B(x)$ for $P_A$- and $P_B$-almost every $x$, then $d=0$ under both endpoint distributions, and the first identity in \eqref{eq:s-aggregate} gives $\Lambda=0$.\\
\indent For part (iv), if $P_A=P_B=P$, the second identity in \eqref{eq:s-aggregate} gives $\Gamma=0$ for any coupling $\pi$. Under condition (ii), exchanging the groups leaves $(P,P)$ unchanged, so $\tau_{\#}\pi=\pi$ for $\tau(x_A,x_B)=(x_B,x_A)$. Hence, with $m^\pi(H):=\mathbb{E}_\pi\{\bar f(X_{\Omega(H)})\}$, we have $m^\pi(H)=m^\pi(H^c)$. For $H\subseteq[p]\setminus\{j\}$, we let $H':=([p]\setminus\{j\})\setminus H$. Since $(H\cup\{j\})^c=H'$, $H^c=H'\cup\{j\}$, and $\kappa_{|H|}=\kappa_{|H'|}$, reindexing \eqref{eq:s-gamma-j} gives $\Gamma_j^\pi=-\Gamma_j^\pi$, so $\Gamma_j^\pi=0$. Under condition (i), $\pi$ is the identity coupling, so $X_{\Omega(H)}=X_A=X_B$ almost surely and $m^\pi(H)=\mathbb{E}_P\{\bar f(X)\}$ for every $H\subseteq[p]$.\\
\indent For part (v), we let tildes denote the quantities after exchanging groups $A$ and $B$. Under condition (ii), transposition equivariance gives $\widetilde\pi=\tau_{\#}\pi$, where $\tau(x_A,x_B)=(x_B,x_A)$. Then $\widetilde d=-d$ and $\widetilde{\bar f}=\bar f$, so \eqref{eq:s-aggregate} gives
\[
\widetilde\Lambda=-\Lambda,
\qquad
\widetilde\Gamma=-\Gamma.
\]
Moreover, under the transposed coupling,
\[
\widetilde X_{\Omega(H)}
=
X_{\Omega(H^c)},
\qquad H\subseteq[p].
\]
For $H\subseteq[p]\setminus\{j\}$, we let $H':=([p]\setminus\{j\})\setminus H$, so that $(H\cup\{j\})^c=H'$ and $H^c=H'\cup\{j\}$. Since $\kappa_{|H|}=\kappa_{|H'|}$, reindexing \eqref{eq:s-gamma-j} gives
\[
\widetilde\Gamma_j^{\,\widetilde\pi}
=
-\Gamma_j^\pi,
\qquad j\in[p],
\]
which proves part (v).
\end{proof}

\section*{S6. Proof of the Reduction to the Blinder--Oaxaca Decomposition}

\mtresult{Proposition 4 (Reduction to the Blinder--Oaxaca decomposition).}
\textit{Suppose that $f_g(x)=\alpha_g+x^\top\beta_g$ for $g\in\{A,B\}$, with $\Delta\alpha$, $\Delta\beta_j$, and $\Delta\mu_{X_j}$ as in the Blinder--Oaxaca decomposition of the main text. Then, for any coupling $\pi$,}
\begin{equation*}
\Lambda
=
\Delta\alpha
+
\sum_{j=1}^p
\frac{\mu_{X_{Aj}}+\mu_{X_{Bj}}}{2}\Delta\beta_j,
\qquad
\Gamma_j^\pi
=
\Delta\mu_{X_j}\frac{\beta_{Aj}+\beta_{Bj}}{2},
\qquad j\in[p].
\end{equation*}
\textit{Hence, with the componentwise Blinder--Oaxaca terms of the main text evaluated at $W=\tfrac12 I_p$, $\Lambda=\Lambda_0^{\mathrm{BO}}+\sum_{j=1}^p\Lambda_j^{\mathrm{BO}}$ and $\Gamma_j^\pi=\Gamma_j^{\mathrm{BO}}$ for every $j\in[p]$.}

\begin{proof}
\indent Since $d(x)=\Delta\alpha+x^\top\Delta\beta$, the first identity in \eqref{eq:s-aggregate} gives the expression for $\Lambda$. The midpoint predictor
\begin{equation*}
\bar f(x)
=
\tfrac12(\alpha_A+\alpha_B)
+
\sum_{j=1}^p
\tfrac12(\beta_{Aj}+\beta_{Bj})x_j
\end{equation*}
is additive, so Proposition~3(ii) with $\psi_j(x_j)=\tfrac12(\beta_{Aj}+\beta_{Bj})x_j$ gives the expression for $\Gamma_j^\pi$ for any coupling $\pi$.\\
\indent At $W=\tfrac12 I_p$, equivalently $w_j=\tfrac12$ for every $j\in[p]$, which is the midpoint reference specification of \citet{reimers1983}, the componentwise Blinder--Oaxaca terms \citep{blinder1973,oaxaca1973} are
\begin{equation*}
\Lambda_0^{\mathrm{BO}}
=
\Delta\alpha,
\qquad
\Lambda_j^{\mathrm{BO}}
=
\frac{\mu_{X_{Aj}}+\mu_{X_{Bj}}}{2}\Delta\beta_j,
\qquad
\Gamma_j^{\mathrm{BO}}
=
\Delta\mu_{X_j}
\left(
\beta_{Bj}+\frac12\Delta\beta_j
\right)
=
\Delta\mu_{X_j}
\frac{\beta_{Aj}+\beta_{Bj}}{2},
\end{equation*}
which proves the second statement.
\end{proof}

\section*{S7. Approximate Allocation and Computational Complexity}

\indent We use the empirical quantities of the Computational Procedures of the main text. For an empirical coupling $\hat\pi\in\mathcal{U}_{n_A,n_B}$, we let
\begin{equation*}
\hat m_{\hat\pi}(H)
:=
\sum_{i=1}^{n_A}
\sum_{i'=1}^{n_B}
\hat\pi_{ii'}\,
\hat{\bar f}\!\left(X_{\Omega(H)}(i,i')\right),
\qquad
H\subseteq[p],
\end{equation*}
denote the internal-game coalition value under $\hat\pi$. Conditional on the data and fitted predictor functions, we let $\hat\Gamma_j$ and $\hat\Gamma_j^{(R)}$ denote the corresponding exact and approximate allocations.\\

\noindent \textbf{Approximate allocation.}
For a uniformly distributed permutation $\rho$ of $[p]$ and $H\subseteq[p]\setminus\{j\}$, $\Pr\{\operatorname{Pre}_j(\rho)=H\}=\kappa_{|H|}$, the weight of $H$ in the exact allocation. Each summand of the approximate allocation $\hat\Gamma_j^{(R)}$ therefore has expectation $\hat\Gamma_j$, so $\hat\Gamma_j^{(R)}$ is unbiased for the exact allocation under permutation sampling for every fixed $R$ \citep{castroetal2009}. Within one permutation, the marginal contributions of the $p$ covariates telescope to $\hat m_{\hat\pi}([p])-\hat m_{\hat\pi}(\emptyset)=\hat\Gamma$, which gives $\sum_{j=1}^p\hat\Gamma_j^{(R)}=\hat\Gamma$.\\

\noindent \textbf{Adaptive permutation budget.}
The budget $R$ may be fixed in advance or selected from pilot permutations. In the adaptive version, we draw $R_0\ge2$ independent uniform pilot permutations $\rho_1^{(0)},\dots,\rho_{R_0}^{(0)}$, let
\begin{equation*}
Z_{mj}^{(0)}
:=
\hat m_{\hat\pi}\!\left(\operatorname{Pre}_j(\rho_m^{(0)})\cup\{j\}\right)
-
\hat m_{\hat\pi}\!\left(\operatorname{Pre}_j(\rho_m^{(0)})\right)
\end{equation*}
denote the marginal contribution of covariate $j$ along pilot permutation $m$, and compute
\begin{equation*}
\widetilde\Gamma_j
:=
\frac{1}{R_0}
\sum_{m=1}^{R_0} Z_{mj}^{(0)},
\qquad
\widetilde\sigma_j^2
:=
\frac{1}{R_0-1}
\sum_{m=1}^{R_0}
\left(Z_{mj}^{(0)}-\widetilde\Gamma_j\right)^2,
\qquad
j\in[p].
\end{equation*}
We measure the Monte Carlo error of the vector of approximate allocations by its relative standard error, $\{\sum_{j}\operatorname{Var}(\hat\Gamma_j^{(R)})\}^{1/2}/\allowbreak(\sum_{j}\hat\Gamma_j^2)^{1/2}$, which for a budget $R$ is estimated from the pilot by $\{\sum_{j}\widetilde\sigma_j^2/(R\sum_{j}\widetilde\Gamma_j^2)\}^{1/2}$. Given a target tolerance $\zeta>0$ and bounds $R_{\min}\le R_{\max}$, we set
\begin{equation*}
R_{\mathrm{req}}
:=
\left\lceil
\frac{\sum_{j=1}^p \widetilde\sigma_j^2}
{\zeta^2\sum_{j=1}^p\widetilde\Gamma_j^2}
\right\rceil,
\qquad
R_{\mathrm{ad}}
:=
\min\!\left\{
R_{\max},
\max\!\left(R_{\min},R_{\mathrm{req}}\right)
\right\},
\end{equation*}
where $R_{\mathrm{req}}$ is the smallest budget at which the estimated relative standard error does not exceed $\zeta$. If $\sum_{j=1}^p\widetilde\Gamma_j^2$ is numerically zero, the relative criterion is undefined, and we set $R_{\mathrm{ad}}:=R_{\min}$ when $\sum_{j=1}^p\widetilde\sigma_j^2$ is also numerically zero and $R_{\mathrm{ad}}:=R_{\max}$ otherwise.\\
\indent The allocation is then computed from $R_{\mathrm{ad}}$ new permutations sampled independently of the pilot, and the pilot contributions are not reused. Conditional on the pilot, $R_{\mathrm{ad}}$ is fixed and the new permutations are independent and uniformly distributed, so the unbiasedness and adding-up properties above hold as for a fixed budget. The criterion targets an estimated relative Monte Carlo precision of the vector of covariate-specific composition effects.\\

\noindent \textbf{Computational complexity.}
We let $L_{\hat\pi}:=\#\{(i,i'):\hat\pi_{ii'}>0\}$ denote the support size of $\hat\pi$ and $C_{\mathrm{pred}}$ the cost of evaluating $\hat{\bar f}$ at one $p$-dimensional covariate vector. For coupling construction, the empirical MSD and entropically regularized couplings require the switching-cost matrix $(C_{ii'})$, whose construction takes $O(n_An_Bp)$ operations and $O(n_An_B)$ storage. When $n_A=n_B=n$, the empirical MSD problem is a linear assignment problem, solvable in $O(n^3)$ operations, and a unique optimal assignment gives $L_{\hat\pi^\star}=n$. The empirical independence coupling requires no optimization and can be represented implicitly in $O(1)$ time and storage. For the empirical entropically regularized coupling, $T$ Sinkhorn iterations require $O(Tn_An_B)$ operations, and the solution has the form $\operatorname{diag}(a)\,(e^{-C_{ii'}/\varepsilon})\,\operatorname{diag}(b)$ with positive vectors $a$ and $b$ \citep{peyrecuturi2019}. Hence $L_{\hat\pi^{\mathrm{ind}}}=L_{\hat\pi^{\varepsilon}}=n_An_B$.\\
\indent For allocation, each coalition value $\hat m_{\hat\pi}(H)$ requires $L_{\hat\pi}$ evaluations of $\hat{\bar f}$. The exact allocation computes and stores all $2^p$ coalition values, requiring $O(2^pL_{\hat\pi}C_{\mathrm{pred}})$ operations for their evaluation, $O(p2^p)$ operations for the Shapley weighting, and $O(2^p)$ storage. With a fixed budget $R$, the approximate allocation evaluates $p+1$ coalition values along each sampled permutation, requiring $O(RpL_{\hat\pi}C_{\mathrm{pred}})$ operations and $O(p)$ storage for the running averages; with an adaptive budget, $R$ is replaced by $R_0+R_{\mathrm{ad}}$. Substituting $L_{\hat\pi^\star}=n$ and $L_{\hat\pi^{\mathrm{ind}}}=L_{\hat\pi^{\varepsilon}}=n^2$ gives the allocation costs in the complexity table of the main text.\\

\section*{S8. Training Specifications of the Predictive Models}

\indent For each group $g\in\{A,B\}$, predictive-model training used a sample of size $n_{\mathrm{train}}=1000$ drawn independently of the sample used for MAD. Tuning was performed within the training sample; unspecified settings used software defaults.\\

\noindent \textbf{Linear regression.}
Linear regression was fitted by least squares with an intercept and all $p$ covariates as main effects, without nonlinear transformations or interaction terms.\\

\noindent \textbf{Gradient boosting machines.}
Gradient boosting was implemented with \texttt{XGBoost} \citep{chenguestrin2016} and \texttt{LightGBM} \citep{keetal2017} using squared-error loss. Maximum depth $\{1,2,3,4\}$ and learning rate $\{0.01,0.05\}$ were selected by 10-fold cross-validation; boosting iterations were selected by early stopping with a maximum of 2,000 rounds and patience 50. The selected configuration was refitted on all $n_{\mathrm{train}}=1000$ observations.\\

\noindent \textbf{Deep neural networks.}
Deep neural networks were implemented with \texttt{Keras} \citep{cholletetal2015} and \texttt{Torch} \citep{paszkeetal2019} using the same training procedure. Both used two ReLU hidden layers, a linear output activation, and mean squared error loss. Hidden-layer widths were selected from $\{(p,\lceil p/2\rceil),(2p,p),(4p,2p)\}$ and dropout from $\{0,0.15,0.3\}$ using two held-out folds of a five-fold partition of the training sample. For each of these two folds, predictors were standardized using the remaining four folds as the fitting subset, and each network was trained with Adam \citep{kingmaba2015} (learning rate $0.01$, batch size $100$) for at most 500 epochs with early stopping on the held-out fold (patience 50) and evaluated at its best validation epoch. The configuration with the smallest mean validation RMSE across the two held-out folds was selected, and the final predictor averaged the predictions of its two best-epoch networks, so that every training observation contributed to the final model without an additional fit.\\

\section*{S9. Population Reference Quantities for the Decomposition Study}

\indent Except for the LA aggregate references given below, the population references are replicate-specific because the coefficient and mean-shift directions and the correlation matrices are regenerated in each decomposition-study replicate. Replicate superscripts are suppressed below. We let $f_g^0=\eta_g^{(\ell)}$ and $\bar f^0=(f_A^0+f_B^0)/2$.\\

\noindent \textbf{Population MSD coupling.}
For the decomposition study,
\begin{equation*}
X_A\sim N_p(\delta,\Sigma_{X_A}),
\qquad
X_B\sim N_p(0,\Sigma_{X_B}),
\end{equation*}
where $\Sigma_{X_A}$ and $\Sigma_{X_B}$ are the unstructured correlation matrices specified in the main text. Because both have unit diagonal, $s_j=1$ in \eqref{eq:s-scale-cost}. The population MSD coupling is the Gaussian quadratic optimal-transport coupling \citep{gelbrich1990,peyrecuturi2019},
\begin{equation*}
X_A
=
\delta+A_{\mathrm{OT}}X_B,
\qquad
A_{\mathrm{OT}}
=
\Sigma_{X_B}^{-1/2}
\bigl(
\Sigma_{X_B}^{1/2}
\Sigma_{X_A}
\Sigma_{X_B}^{1/2}
\bigr)^{1/2}
\Sigma_{X_B}^{-1/2}.
\end{equation*}
Hence every hybrid vector $X_{\Omega(H)}$ is Gaussian.\\

\noindent \textbf{Aggregate reference quantities.}
With $\mu_{r\mid s}^0:=\mathbb{E}_{P_s}\{f_r^0(X)\}$ for $r,s\in\{A,B\}$,
\begin{equation*}
\Delta^0
=
\mu_{A\mid A}^0-\mu_{B\mid B}^0,
\qquad
\Lambda^0
=
\tfrac12\bigl(\mu_{A\mid A}^0+\mu_{A\mid B}^0-\mu_{B\mid A}^0-\mu_{B\mid B}^0\bigr),
\end{equation*}
\begin{equation*}
\Gamma^0
=
\tfrac12\bigl(\mu_{A\mid A}^0+\mu_{B\mid A}^0-\mu_{A\mid B}^0-\mu_{B\mid B}^0\bigr),
\qquad
\Delta^0=\Lambda^0+\Gamma^0.
\end{equation*}
We write $\mu_{X_A}=\delta$ and $\mu_{X_B}=0$, let $K=2p/5$, and define
\begin{equation*}
q_{g\mid s,j}
:=
\mathbb{E}_{P_s}\{\phi_{r_g(j)}(X_j)\},
\qquad
m_{g\mid s,k}
:=
\mathbb{E}_{P_s}\{\chi_{g,k}(X)\},
\end{equation*}
for $g,s\in\{A,B\}$, $j\in[p]$, and $k=1,\ldots,K$, and let $m_{g\mid s}=(m_{g\mid s,1},\ldots,m_{g\mid s,K})^\top$. Then
\begin{align*}
\mu_{A\mid s}^{0,\mathrm{LA}}
&=
\alpha+(\beta+\gamma)^\top\mu_{X_s},
&
\mu_{B\mid s}^{0,\mathrm{LA}}
&=
\beta^\top\mu_{X_s},\\
\mu_{A\mid s}^{0,\mathrm{LM}}
&=
\alpha+(\beta+\gamma)^\top\mu_{X_s}
+(\theta+\xi)^\top m_{A\mid s},
&
\mu_{B\mid s}^{0,\mathrm{LM}}
&=
\beta^\top\mu_{X_s}+\theta^\top m_{B\mid s},\\
\mu_{A\mid s}^{0,\mathrm{NA}}
&=
\alpha+\sum_{j=1}^p(\beta_j+\gamma_j)q_{A\mid s,j},
&
\mu_{B\mid s}^{0,\mathrm{NA}}
&=
\sum_{j=1}^p\beta_jq_{B\mid s,j},\\
\mu_{A\mid s}^{0,\mathrm{NM}}
&=
\alpha+\sum_{j=1}^p(\beta_j+\gamma_j)q_{A\mid s,j}
+(\theta+\xi)^\top m_{A\mid s},
&
\mu_{B\mid s}^{0,\mathrm{NM}}
&=
\sum_{j=1}^p\beta_jq_{B\mid s,j}
+\theta^\top m_{B\mid s}.
\end{align*}
Because every marginal under $P_B$ is $N(0,1)$ and each $\phi_k$ is centered to have mean zero under $N(0,1)$, $q_{A\mid B,j}=q_{B\mid B,j}=0$ for all $j$. The interaction expectations are obtained from Gaussian second- and third-order moments \citep{isserlis1918}. In particular, if $X\sim N_p(\mu,\Sigma)$,
\begin{equation*}
\mathbb{E}(X_aX_b)=\mu_a\mu_b+\Sigma_{ab},
\end{equation*}
and
\begin{equation*}
\mathbb{E}(X_aX_bX_c)
=
\mu_a\mu_b\mu_c
+\mu_a\Sigma_{bc}
+\mu_b\Sigma_{ac}
+\mu_c\Sigma_{ab}.
\end{equation*}
The nonlinear expectations for $\phi_1$, $\phi_2$, and $\phi_4$ are evaluated in closed form, whereas $\phi_3$ is evaluated by one-dimensional numerical integration.\\
\indent For LA,
\begin{equation*}
\beta^\top\delta=\gamma^\top\delta=\frac{p}{4},
\end{equation*}
so
\begin{equation*}
(\mu_{A\mid A}^0,\mu_{A\mid B}^0,\mu_{B\mid A}^0,\mu_{B\mid B}^0)
=
\left(
1+\frac{p}{2},
1,
\frac{p}{4},
0
\right),
\end{equation*}
and
\begin{equation*}
(\Delta^0,\Lambda^0,\Gamma^0)
=
\left(
1+\frac{p}{2},
1+\frac{p}{8},
\frac{3p}{8}
\right).
\end{equation*}
For LM, NA, and NM, the references are evaluated replicate by replicate from the formulas above.\\

\noindent \textbf{Covariate-specific reference quantities.}
The reference $\Gamma_j^0$ is $\Gamma_j^{\pi^\star}$ in \eqref{eq:s-gamma-j} evaluated with $\bar f^0$, where $\pi^\star$ is the population MSD coupling above. For LA and NA, $\bar f^0(x)=\alpha/2+\sum_{j=1}^p\psi_j(x_j)$ is additive, with $\psi_j(t)=(\beta_j+\gamma_j/2)\,t$ for LA and $\psi_j(t)=\tfrac12\{(\beta_j+\gamma_j)\phi_{r_A(j)}(t)+\beta_j\phi_{r_B(j)}(t)\}$ for NA, so Proposition~3(ii) gives, for LA,
\begin{equation*}
\Gamma_j^0
=
\left(\beta_j+\frac{\gamma_j}{2}\right)\delta_j
=
\frac{p}{4}
\left(\nu_{1j}+\frac12\nu_{2j}\right)
(\nu_{1j}+\nu_{2j}+\nu_{3j}),
\end{equation*}
with
\begin{equation*}
\sum_{j=1}^p\Gamma_j^0
=
\beta^\top\delta+\frac12\gamma^\top\delta
=
\frac{3p}{8}
=
\Gamma^0,
\end{equation*}
and, for NA,
\begin{align*}
\Gamma_j^0
&=
\frac{\beta_j+\gamma_j}{2}
\left[
\mathbb{E}_{P_A}\{\phi_{r_A(j)}(X_j)\}
-
\mathbb{E}_{P_B}\{\phi_{r_A(j)}(X_j)\}
\right]\\
&\quad+
\frac{\beta_j}{2}
\left[
\mathbb{E}_{P_A}\{\phi_{r_B(j)}(X_j)\}
-
\mathbb{E}_{P_B}\{\phi_{r_B(j)}(X_j)\}
\right].
\end{align*}
For LM and NM, the midpoint predictors are
\begin{equation*}
\bar f_{\mathrm{LM}}^0(x)
=
\frac{\alpha}{2}
+
\sum_{j=1}^p
\left(\beta_j+\frac{\gamma_j}{2}\right)x_j
+
\frac12\sum_{k=1}^K
\left[
(\theta_k+\xi_k)\chi_{A,k}(x)
+
\theta_k\chi_{B,k}(x)
\right],
\end{equation*}
and
\begin{align*}
\bar f_{\mathrm{NM}}^0(x)
&=
\frac{\alpha}{2}
+
\frac12\sum_{j=1}^p
\left[
(\beta_j+\gamma_j)\phi_{r_A(j)}(x_j)
+
\beta_j\phi_{r_B(j)}(x_j)
\right]\\
&\quad+
\frac12\sum_{k=1}^K
\left[
(\theta_k+\xi_k)\chi_{A,k}(x)
+
\theta_k\chi_{B,k}(x)
\right].
\end{align*}
Substitution into \eqref{eq:s-gamma-j} reduces the interaction contributions to second- and third-order moments of the Gaussian hybrid vectors. These moments are evaluated analytically under the population MSD coupling for each replicate.\\

\section*{S10. Coalition Deviation under Approximate Allocation}

\indent In the coupling study, the coalition deviation of an empirical coupling $\hat\pi$ is
\begin{equation*}
D_{\hat\pi}
:=
\Bigl\{
2^{-p}\sum_{H\subseteq[p]}
\bigl[\hat m_{\hat\pi}(H)-\hat m_{\hat\pi}(\emptyset)\bigr]^2
\Bigr\}^{1/2},
\end{equation*}
with $\hat m_{\hat\pi}(H)$ as in Section~S7. For $p\in\{5,10,15\}$, where the exact allocation is used, $D_{\hat\pi}$ is computed from all $2^p$ coalition values. For $p=20$, where the approximate allocation is used, it is estimated from the same sampled permutations $\rho_1,\dots,\rho_R$. With $H_{mk}$ denoting the set of the first $k$ covariates in $\rho_m$, for $k=0,\dots,p$, we use
\begin{equation*}
\widehat D_{\hat\pi,R}
:=
\left[
\sum_{k=0}^p
\frac{\binom{p}{k}}{2^p}\,
\frac{1}{R}
\sum_{m=1}^R
\left\{
\hat m_{\hat\pi}(H_{mk})
-
\hat m_{\hat\pi}(\emptyset)
\right\}^2
\right]^{1/2}.
\end{equation*}
For each $k$, $H_{mk}$ is uniform over subsets of size $k$, so $\widehat D_{\hat\pi,R}^2$ is unbiased for $D_{\hat\pi}^2$. The calculation reuses the allocation paths and requires no additional coalition evaluations.\\

\section*{S11. Details of the Real Data Application}

\noindent \textbf{Data processing.}
We used the closed-reference operational taxonomic unit table of the AGP \citep{mcdonaldetal2018}, in which reads trimmed to 100 nucleotides were clustered at 97\% similarity against the Greengenes 13\_8 reference \citep{desantisetal2006,mcdonaldetal2012}, without rarefaction. Bacterial counts were aggregated to the genus level after bracketed genus names were merged with their unbracketed counterparts and reads without a genus assignment were removed. We retained fecal samples from US and UK residents aged 20--70 years with a recorded sex, a BMI of at least 15 and below 60, and at least 10,000 reads, and kept the sample with the largest number of reads for each participant. Genera with a mean relative abundance of at least 0.001 among the resulting 6,673 participants were retained, and the analysis was then restricted to males aged 30--70 years. Zero counts were replaced by 0.5 before the CLR transformation.\\

\noindent \textbf{Training procedures.}
Each predictive model was fitted separately to each country sample, with BMI as the response and the $p=53$ CLR coordinates as covariates; unspecified settings used software defaults. Linear regression was fitted by least squares with an intercept and all CLR coordinates as main effects; because the CLR coordinates sum to zero, its coefficients are determined only up to a common additive constant, which leaves the fitted predictor on the CLR hyperplane unchanged. Gradient boosting with \texttt{XGBoost} \citep{chenguestrin2016} and \texttt{LightGBM} \citep{keetal2017} used the squared-error loss; maximum depth $\{1,2,3,4\}$ and learning rate $\eta\in\{0.01,0.05\}$ were selected by two repeats of 10-fold cross-validation, boosting iterations were selected by early stopping with a maximum of 2,000 rounds and patience 50 at $\eta=0.05$ and 10,000 rounds and patience 250 at $\eta=0.01$, and the selected configuration was refitted on all observations. Deep neural networks with \texttt{Keras} \citep{cholletetal2015} and \texttt{Torch} \citep{paszkeetal2019} used two ReLU hidden layers, a linear output activation, and the mean squared error loss. Hidden-layer widths $\{(\lceil p/2\rceil,\lceil p/4\rceil),(p,\lceil p/2\rceil),(2p,p)\}$ and dropout $\{0,0.1,0.2,0.3\}$ were selected by five-fold cross-validation. Within each fold, predictors and BMI were standardized using the fitting subset, and each network was trained with Adam \citep{kingmaba2015} (learning rate $0.01$, batch size $100$) for at most 500 epochs with early stopping on the held-out fold (patience 50) and retained at its best validation epoch; its predictions were transformed back to the BMI scale and shifted to match the mean BMI of the fitting subset. The final predictor averaged the five networks of the selected configuration and was shifted to match the mean BMI of the country sample.\\

\noindent \textbf{Predictive performance.}
Predictive performance was assessed by the out-of-fold root mean squared error (RMSE) in 10-fold cross-validation within each country, with all tuning carried out within the training folds, and was averaged over the two countries. The average RMSEs (kg/m$^2$) of linear regression, \texttt{XGBoost}, \texttt{LightGBM}, \texttt{Keras}, and \texttt{Torch} were 4.179, 4.189, 4.161, 4.180, and 4.162, respectively, and \texttt{LightGBM}, with the smallest average RMSE, was selected. Applying its tuning procedure to each full country sample selected trees of depth one with $\eta=0.05$ and 365 (UK) and 314 (US) boosting iterations.\\

\noindent \textbf{Evaluation at hybrid covariate vectors.}
We let $\Pi_{\mathrm{CLR}}:=I_p-p^{-1}\mathbf{1}_p\mathbf{1}_p^\top$ denote the orthogonal projection onto the CLR hyperplane $\{x\in\mathbb{R}^p:\mathbf{1}_p^\top x=0\}$, and we applied MAD to $\tilde f_g(x):=\hat f_g(\Pi_{\mathrm{CLR}}x)$ for $g\in\{A,B\}$. Because $\Pi_{\mathrm{CLR}}x=x$ for every observed CLR vector, $\tilde f_g$ and $\hat f_g$ coincide on both samples, so that $\Delta$, $\Lambda$, and $\Gamma$ are unchanged and only the internal-game coalition values at intermediate hybrid covariate vectors are affected.\\

\noindent \textbf{Coupling and allocation.}
The empirical MSD coupling was supported on $n_A+n_B-1=2{,}593$ pairs. The adaptive budget of Section~S7 used $R_0=500$, $R_{\min}=500$, $R_{\max}=100{,}000$, and $\zeta=0.0045$, slightly below the target of 0.005, which gave $R_{\mathrm{ad}}=646$ and an achieved relative Monte Carlo standard error of 0.0045. Table~\ref{tab:s-agp-gamma} lists the genus-specific composition effects of all 53 genera; their Monte Carlo standard errors were at most $1.1\times10^{-4}$.\\

\begin{table}[ht]
\centering
\caption{Genus-specific composition effects $\Gamma_j$ (kg/m$^2$, UK minus US) of the 53 genera, in decreasing order of $|\Gamma_j|$ down the left and then the right column; $\sum_{j=1}^{53}\Gamma_j=\Gamma=-0.100$.}
\label{tab:s-agp-gamma}
\begin{tabular}{lrlr}
\toprule
Genus & $\Gamma_j$ & Genus & $\Gamma_j$ \\
\midrule
\textit{Corynebacterium} & 0.0765 & \textit{Bacillus} & 0.0028 \\
\textit{Finegoldia} & $-$0.0648 & \textit{Megasphaera} & $-$0.0021 \\
\textit{Bifidobacterium} & 0.0439 & \textit{Turicibacter} & 0.0021 \\
\textit{Clostridium} & $-$0.0345 & \textit{Stenotrophomonas} & $-$0.0020 \\
\textit{Haemophilus} & $-$0.0281 & \textit{Eubacterium} & $-$0.0018 \\
\textit{Akkermansia} & $-$0.0214 & \textit{Desulfovibrio} & 0.0016 \\
\textit{Butyricimonas} & 0.0202 & \textit{Anaerococcus} & $-$0.0016 \\
\textit{Dorea} & $-$0.0154 & \textit{Enterobacter} & 0.0015 \\
\textit{Collinsella} & $-$0.0151 & \textit{Serratia} & 0.0015 \\
\textit{Catenibacterium} & 0.0109 & \textit{Lachnospira} & $-$0.0012 \\
\textit{Ruminococcus} & $-$0.0096 & \textit{Paraprevotella} & 0.0012 \\
\textit{Prevotella} & $-$0.0082 & \textit{Streptococcus} & $-$0.0011 \\
\textit{Blautia} & $-$0.0082 & \textit{Morganella} & $-$0.0010 \\
\textit{Lachnobacterium} & $-$0.0078 & \textit{Bilophila} & $-$0.0006 \\
\textit{Peptoniphilus} & $-$0.0063 & \textit{Staphylococcus} & $-$0.0006 \\
\textit{Porphyromonas} & $-$0.0063 & \textit{Fusobacterium} & $-$0.0006 \\
\textit{Dialister} & $-$0.0061 & \textit{Sutterella} & 0.0005 \\
\textit{Roseburia} & $-$0.0060 & \textit{Coprococcus} & $-$0.0004 \\
\textit{Anaerostipes} & $-$0.0055 & \textit{Pseudomonas} & $-$0.0003 \\
\textit{Odoribacter} & 0.0045 & \textit{Bacteroides} & 0.0003 \\
\textit{Lactobacillus} & $-$0.0042 & \textit{Veillonella} & $-$0.0003 \\
\textit{Listeria} & $-$0.0041 & WAL\_1855D & 0.0002 \\
\textit{Succinivibrio} & 0.0036 & \textit{Acinetobacter} & 0.0002 \\
\textit{Parabacteroides} & $-$0.0035 & \textit{Brevundimonas} & $-$0.0001 \\
\textit{Faecalibacterium} & 0.0033 & \textit{Comamonas} & 0.0001 \\
\textit{Oscillospira} & $-$0.0032 & \textit{Proteus} & $-$0.0001 \\
\textit{Phascolarctobacterium} & $-$0.0030 &  &  \\
\bottomrule
\end{tabular}
\end{table}
\FloatBarrier